\documentclass[lettersize,journal]{IEEEtran}
\usepackage{amsmath,amsfonts}
\usepackage{algorithmic}
\usepackage{array}
\usepackage{textcomp}
\usepackage{stfloats}
\usepackage{url}
\usepackage{graphicx}
\usepackage[font=normal]{caption}
\usepackage{pifont} 
\usepackage{makecell}  
\usepackage{subcaption}
\usepackage{nccmath}
\usepackage{multirow} 
\usepackage{textcomp}
\usepackage[ruled,vlined]{algorithm2e}
\usepackage{adjustbox}
\usepackage{tikz}
\usepackage{pgfplots}
\pgfplotsset{compat=1.18}
\usepackage{xcolor}
\usepackage{pgfplotstable}
\usepackage{nomencl}
\usepackage{booktabs}

\begin{document}

\title{Communication-Guided Multi-Mutation Differential Evolution for Crop Model Calibration}

\author{Sakshi Aggarwal, Mudasir Ganaie, Mukesh Saini}


\markboth{Journal of \LaTeX\ Class Files,~Vol.~14, No.~8, August~2021}%
{Shell \MakeLowercase{\textit{et al.}}: A Sample Article Using IEEEtran.cls for IEEE Journals}

\IEEEpubid{0000--0000/00\$00.00~\copyright~2021 IEEE}

\maketitle
\makenomenclature
\begin{abstract}
In this paper, we propose a multi-mutation optimization algorithm, Differential Evolution with Multi-Mutation Operator-Guided Communication (DE-MMOGC), implemented to improve the performance and convergence abilities of standard differential evolution in uncertain environments. DE-MMOGC introduces a communication-guided scheme integrated with multiple mutation operators to encourage exploration and avoid premature convergence. Along with this, it includes a dynamic operator selection mechanism to use the best-performing operator over successive generations. To assimilate real-world uncertainties and missing observations into the predictive model, the proposed algorithm is combined with the Ensemble Kalman Filter. To evaluate the efficacy of the proposed DE-MMOGC in uncertain systems, the unified framework is applied to improve the predictive accuracy of crop simulation models. These simulation models are essential to precision agriculture, as they make it easier to estimate crop growth in a variety of unpredictable weather scenarios. Additionally, precisely calibrating these models raises a challenge due to missing observations. Hence, the simplified WOFOST crop simulation model is incorporated in this study for leaf area index (LAI)-based crop yield estimation. DE-MMOGC enhances the WOFOST performance by optimizing crucial weather parameters (temperature and rainfall), since these parameters are highly uncertain across different crop varieties, such as wheat, rice, and cotton. The experimental study shows that DE-MMOGC outperforms the traditional evolutionary optimizers and achieves better correlation with real LAI values. We found that DE-MMOGC is a resilient solution for crop monitoring.
\end{abstract}


\nomenclature[a]{$a_t$}{State variable at time t in EnKF}
\nomenclature[c1]{C(t)}{Predicted LAI at time t by WOFOST}
\nomenclature[c2]{Cr}{Crossover}
\nomenclature[f1]{F}{Scaling Factor}
\nomenclature[f2]{F(x)}{Minimum objective function}
\nomenclature[g]{G}{Generations}
\nomenclature[k1]{K(t)}{Vector of climate conditions at time t}
\nomenclature[k2]{$K_t$}{Kalman Gain}
\nomenclature[m]{M}{Ensemble size}
\nomenclature[n]{np}{Number of solutions in a population}
\nomenclature[p]{$P_t$}{Covariance matrix computed for time t}
\nomenclature[q]{$Q_t$}{Covariance for process noise}
\nomenclature[r1]{RAIN}{Rain parameter to be optimized by DE-MMOGC}
\nomenclature[r2]{$R_t$}{Covariance for observation noise}
\nomenclature[s]{S}{Set of mutant operators}
\nomenclature[t]{TAVG}{Temperature parameter to be optimized by DE-MMOGC}
\nomenclature[u]{U}{Set of trial solutions}
\nomenclature[v]{V}{Set of mutant solutions}
\nomenclature[x1]{X}{Population}
\nomenclature[x2]{x}{Individual solution}
\nomenclature[y]{$y_t$}{Perturbed observation state in EnKF}

\section{Introduction} 
\label{sec: introduction}

Differential evolution (DE) \cite{Storn1997} is an effective evolutionary method that is well known for its simplicity and resilience in solving global optimization problems. In agricultural sciences, crop simulation model optimization is crucial for increasing resource utilization, sustainability, and productivity \cite{WANG2024108899}. Because traditional DE approaches rely on single-mutation procedures, their exploration capabilities may be restricted in hard optimization environments \cite{Elsayed2012}. To overcome these constraints, a multi-mutation operator can be used to greatly increase the range of solutions generated during the optimization process \cite{xmode, KMSallam2019,Elsayed2017,sallam2018,sallam2020}. The use of multi-mutation approaches in DE can further increase crop simulation model performance and lead to long-term benefits in agricultural practice and decision-making. Using multi-mutation operators, this research proposes a variant of differential evolution that can improve crop simulation models and support sustainable agricultural development.

\par The growing demand for sustainable agricultural output around the world requires sophisticated methods for accurate crop growth forecasting and resource management. Crop simulation models (CSMs), such as DSSAT \cite{dssat}, APSIM \cite{apsim}, and WOFOST \cite{WOFOST}, are a crucial part of this endeavour. These methods incorporate biophysical processes to mimic crop growth under different weather and management conditions. The prediction performance of CSM is enhanced by techniques like genetic algorithm \cite{10.1145/3727993.3728022} and differential evolution \cite{11146161}. Data assimilation techniques \cite{10.1145/2379810.2379816, ZHANG2015291}, which can address low-quality observations and ambiguous data-gathering criteria, have received the most attention among such methodologies. For these models to be accurate, a various parameters need to be adjusted, ranging from phenological features to soil-plant relations. The inability of conventional optimization methods, such as gradient-based algorithms \cite{TIWARI202041}, to manage this complexity frequently results in less-than-ideal solutions or even convergence failure. 

\par Crop growth models are tailored for weather stress factors in order to forecast yields, enhance farming methods, and reduce weather-related risks \cite{TALBOT202438, ECHCHATIR2025109548, 10.1002}. Determining crop varieties that may produce larger and more consistent yields under particular agro climatic conditions requires calibrating average temperature and precipitation parameters \cite{Mankin2024, WEBBER2025110032}. While precipitation controls soil moisture availability and the degree of drought stress, temperature controls vital physiological processes like germination, phenological development, flowering, and grain filling \cite{Akhavizadegan2021}. Greater susceptibility to heat stress or underestimation of water deficits are a few examples of how even small biases in these environmental factors can lead to significant inaccuracies in simulating varietal performance. Crop simulation models can better represent varietal differences in growth duration, stress tolerance, and yield potential by accurately calibrating temperature and precipitation inputs. This allows for the selection of crop varieties that are more resilient to future climate variability and better adapted to local environments \cite{ECHCHATIR2025109548}.


\par To address the limitations of underlined CSMs, we apply our proposed DE-MMOGC to handle the environmental stress factors and improve the CSM projections. We include the modified WOFOST simulation model in this study. We propose an optimizer that can improve the crop simulation model. The influencing weather factors are shown in Table S1 of the Supplementary Material. We consider the weather parameters \textit{TAVG} and \textit{RAIN} for a basic flow while maintaining rest as constant. Furthermore, a data assimilation method is also used to simulate uncertain and missing observations.
\clearpage
The key contributions are highlighted below:

\begin{enumerate}
    \item This research proposes a new optimization algorithm, DE-MMOGC, using multiple mutation operators of differential evolution. DE-MMOGC uses dynamic operator selection and communication medium for improving the performance of the operators over the generations.
    

    \item The proposed method is applied to improve crop simulation models by optimizing weather parameters. In order to increase the CSM's predictive ability, we strive to optimize its weather inputs.
\end{enumerate}

Section S2 (Background) in the Supplementary Material gives the comprehensive background of the concepts implemented in our research. The experimental setup with data, parameter settings, and evaluation metrics is provided in Section S3 (Experimental Setup) in the Supplementary Material.

\printnomenclature
\section{Literature}
\label{literature}

Several optimization algorithms have been applied to model calibration, but this literature draws attention towards DE strategies in improving calibration \cite{Zhang2025, 10.3389/fpls.2025.1617775, Zhao2025}. In this section, first, existing studies on model calibration using evolutionary algorithms and DE are discussed. In the subsequent section, research pertaining to multi-operator DE variants is explored. The CSMs are discussed in Section S1 (Crop Simulation Models) of the Supplementary Material with their limitations.    

\subsection{Model Calibration}
Model calibration \cite{10.1162/evco_a_00325, Zhang2025} constitutes a pivotal challenge in the domains of computational modeling, wherein the objective is to ascertain the most appropriate configuration of parameters that facilitates the maximal performance of a system. Zhang et al. (2025) \cite{Zhang2025} proposed a Bayesian–evolutionary hybrid framework for calibrating nonlinear simulation models after studying evolutionary-based model calibration under climatic uncertainty. Although the method showed enhanced robustness on several datasets, examining its applicability to high-dimensional parameter spaces constitutes an important avenue for future research. Nicușan et al. (2025)  \cite{turn0academia18} proposed ACCES, an evolutionary calibration framework that uses meta-programming and non-invasive optimization for complicated simulators, to address efficiency difficulties in large-scale model calibration. However, future studies may focus on assessing the scalability of these methods in the presence of substantial observational noise. Tangherloni et al. (2024) \cite{10611920} focused on real-parameter fine-tuning under uncertain constraints. The research proposed a modified version of the Enhanced Adaptive Cooperative Particle Swarm Optimization (EACOP) algorithm, aimed at real-parameter single-objective optimization. The consistency of performance improvements across diverse problem domains continues to be a subject of active discussion. To fine-tune deep neural networks at the subspace level, Whitaker and Whitley (2023) \cite{10.1145/3583133.3590705} suggested Sparse Mutation Decompositions (SMD) for neuro-evolution. The design of sparse operators has not been extensively studied in previous neuro-evolution research. Coevolutionary techniques to warm-start AlphaZero's self-play reinforcement learning were investigated by Wang et al. (2020) \cite{wang2020warm}. The applicability of the findings may be further broadened by reducing the reliance on AlphaZero’s specific architectural choices and reward formulation.

\par Apart from the above techniques, DE is also adopted for parameter adjustments in various domains. Salehinejad et al.(2024) \cite{salehinejad2025efficacy} employed DE in conjunction with local search methodologies to enhance the optimization of connection weights within artificial neural networks. Their hybrid framework demonstrated superior generalization and convergence metrics when contrasted with conventional backpropagation techniques, particularly in scenarios characterized by intricate error landscapes. A thorough calibration methodology tailored for soil-crop models was proposed by Wallach et al. (2024) \cite{turn0search1}, emphasizing the significance of simultaneous multi-variable fitting and statistical parameter selection for reliable calibration results. Iterative Model Calibration (IMC) and other recent work on automatic calibration frameworks provide scalable, gradient-free techniques that can handle vast parameter spaces in numerical models without requiring derivative information \cite{turn0search17}. Studies on hydrological and distributed models, which go beyond agricultural systems, further support the benefit of DE and other EAs: multi-objective regional calibration with decision tree-based techniques improves spatial consistency of parameters across vast catchments \cite{turn0search0}. Multi-objective versions of DE, such as multi-operator DE and non-dominated sorting \cite{aggarwal2022multi,peng2023multi}, have been developed to balance several competing goals and adjust parameter settings.

\subsection{Multi-mutation DE}

The majority of research on parameter optimization with evolutionary algorithms has concentrated on benchmark comparisons or discrete algorithmic enhancements.

Sallam and Chakrabortty (2020) \cite{sallam2020evolutionary} presented a two-stage evolutionary strategy that enhances exploration and exploitation capabilities by utilizing multiple mutation operators. The study highlighted how choosing the right operator can increase the robustness of the algorithm. The authors Biswas, Abbasi, and Das (2021) \cite{biswas2022two} provided a hybrid approach to restricted optimization problems that combines multi-operator DE and VIKOR decision-making. In comparison to traditional DE variations, their empirical results showed notable performance increases. Through the use of improved control parameter adaptation techniques across several operators, the research of Yuan et al. (2024) \cite{yuan2025improved} delivered an improved version of DE. The work helped solve problems in the real world, where resilience and flexibility are essential. In the recent research, Reda et al. (2025) \cite{reda2025novel}, multi-operator DE and reinforcement learning are integrated in a unique way to dynamically pick operators based on feedback from the environment. An important step forward in managing dynamic and time-variant optimization problems is represented by this adaptive system.
\par In multi-mutation DE, operator selection is crucial because different mutation techniques show differing strengths in different problem landscapes and evolutionary stages. Khan et al. (2024) \cite{Khan2024AOSDE} proposed an adaptive operator selection framework that exhibits better robustness than fixed single-operator DE variants, giving higher selection probabilities to mutation methods that produce superior offspring. Despite their effectiveness, these techniques mostly depend on carefully thought-out credit assignment procedures, which could result in extra computing costs. Learning-based operator selection algorithms have become a viable substitute in recent times. In order for the algorithm to learn context-dependent operator preferences, Ma et al. (2025) \cite{Wang2024RLDE} used reinforcement learning to simultaneously change control parameters and mutation methods. Higher computing costs and sensitivity to learning hyperparameters limit the benefits of these approaches, despite their improved adaptability and performance on challenging benchmarks. In contrast, 
cooperation amongst mutation operators is highlighted in multi-operator DE systems with strategy sharing or migration. In order to achieve faster convergence than independent operator selection schemes, Gong et al. (2024)  \cite{Gong2024MultiOperatorDE} developed a strategy-sharing mechanism that enables operators to share successful search information. 
\par While learning-based approaches provide better adaptation, subpopulation-based strategies provide diversity, and adaptive operator selection methods provide flexibility. The necessity for hybrid or multi-operator DE frameworks that can dynamically exploit the capabilities of various operator selection procedures is highlighted by the fact that no single mechanism consistently outperforms others across all problem domains \cite{Qin2009SaDE}.


 
\section{Problem Formulation for CSM}
\label{formulation}

Let $K(t)$ represent a vector of climate conditions at time $t$. These climatic measurements are soil specifications, crop characteristics, and weather inputs. The WOFOST model can simulate the LAI, under climatic conditions $K(t)$ at time $t+1$, in the following way:
\begin{flalign}
\label{eq:prediction}
        &C(t+1) = WOFOST (C(0), C(1), C(2),...,C(t),K(t))&
\end{flalign}
where $C(t+1)$ denotes the predicted LAI at time $(t+1)$ by WOFOST. Initially, $t=0$, $C(0)=0$. 

\par In this research, soil and crop parameters corresponding to the crop variety are assumed to be fixed, whereas the weather inputs are thought to be inconsistent. Therefore, the vector $K(t)$ can be illustrated as, 
\begin{flalign}
    \label{eq:vector}
        &K(t) = \begin{bmatrix}
            K_1(t)\\
            K_2
        \end{bmatrix}&
\end{flalign}    
Here, $ K_1(t) = \begin{bmatrix}
    TAVG\\ RAIN
\end{bmatrix} $ and the rest of parameters are stored in $K_2  =  \begin{bmatrix}
    Soil & Crop & IRRAD
\end{bmatrix}^T$

To replicate the uncertainties of crop and soil specifications, a random Gaussian noise is enumerated with distribution $\mathcal{N}(0,0.1)$. Hence, the equation (\ref{eq:vector}) can be rewritten with the perturbation $\epsilon_i$ as
\begin{flalign}
  &K(\epsilon_i,t) = \begin{bmatrix}
            K_1(t)\\
            K_2+\epsilon_i 
        \end{bmatrix}&  
\end{flalign}
However, the equation (\ref{eq:prediction}) for predicting LAI is modified as
    \begin{flalign}  
    \begin{split}
      C(\epsilon_i, t+1) = WOFOST (C(0), C(\epsilon_1,1), C(\epsilon_2, 2),..., \\  C(\epsilon_i,t),K(\epsilon_i,t)) 
    \end{split}       
    \end{flalign}
Finally, for $M$ ensemble size, the predicted assimilated LAI could be modeled by
    \begin{flalign}
       &C(t+1) = \frac{\sum_{i=1}^M C(\epsilon_i, t+1)}{M}&
    \end{flalign}
Typically, to improve the predictive performance of CSM, we underline mean square error (MSE) as the objective function between true LAI and assimilated LAI. The aim is to \textit{minimize} the error between true values and predicted values. With every iteration, it tries to bridge the gap between target LAI and estimated LAI corresponding to a variety of particular crops. Section \ref{sec:calibration} covers the parameter calibration and objective function in DE-MMOGC for improved CSM in depth.

\par Say, $X= [x_1,x_2,x_3,...,x_{np}]$ is a $np$-sized parameter vector to be optimized under the constraint $x^{lower}\leq x_i \leq x^{upper}$ bounds. The objective function is $F(x)$ defined as a minimum function for a given set of parameters $X$. $y_i$ denotes the true LAI value. The estimated LAI value is $\hat{y}_i = C(t+1)_i$ by the predictive model. Therefore, given $np$ simulated days, MSE is defined as
\begin{flalign}
    \label{MSE}
        &MSE = \frac{1}{np} \sum_{i=1}^{np} (y_i- C(t+1)_i)^2&\\
        &MSE = \frac{1}{np} \sum_{i=1}^{np} (y_i- \hat{y}_i)^2&
\end{flalign}
It becomes objective function in the proposed optimizer DE-MMOGC as 
    \begin{flalign}
        &Minimize\;: F(x) = MSE&
    \end{flalign} 

\section{Proposed Optimization Algorithm— DE-MMOGC}
\label{proposed}

\begin{figure*}
    \centering
    \usetikzlibrary{arrows.meta, positioning, shapes.geometric, fit}

   \tikzset{
  process/.style = {rectangle, rounded corners=3pt, draw=black, thick, align=center,
                    minimum width=3cm, minimum height=1cm, font=\small, fill=blue!10},
  subpop/.style = {rectangle, rounded corners=3pt, draw=black, thick, align=center,
                   minimum width=3cm, minimum height=1cm, font=\small, fill=gray!10},
  dashedbox/.style = {draw=black, thick, dashed, rounded corners=3pt, inner sep=3pt},
  arrow/.style = {->, thick, >=Stealth, draw=blue!70!black},
  labeltext/.style = {font=\small}
}
    \begin{tikzpicture}[node distance=1cm]

\node[process, fill=green!15] (init) {Population};
\node[labeltext, right =0.2cm of init.east] {Population Initialization};

\node[subpop, below=of init, xshift=-3.4cm] (sub1) {Subpopulation 1:\\DE/current-to-best/1};
\node[subpop, below=of init] (sub2) {Subpopulation 2:\\DE/rand-to-best/2};
\node[subpop, below=of init, xshift=3.4cm] (sub3) {Subpopulation 3:\\DE/current-to-pbest/1};

\node[dashedbox, fit=(sub1)(sub2)(sub3)] (mutationbox) {};
\node[labeltext, right =0.2cm of mutationbox.east] {Multi-Mutation Operators};

\node[process, below=1.0cm of mutationbox, fill=yellow!20, xshift=-3.1cm] (div1) {Find the worst solutions \\ from each subpopulation};
\node[process, below=1.0cm of mutationbox, fill=yellow!20, xshift=3cm] (div2) {Find $k\%$ best solutions \\ from the entire population};
\node[dashedbox, fit=(div1)(div2)] (diversitybox) {};
\node[labeltext, right=0.1cm of diversitybox.east] {Diversity and Convergence};

\node[process, below=1cm of diversitybox, fill=red!15] (comm) {
  Replace the worst solutions of each subpopulation \\
  with $k\%$ of the best solutions within the entire population
};
\node[labeltext, right=0.1cm of comm.east] {Communication Guided Scheme};

\node[subpop, below=of comm, xshift=-3.5cm] (dyn1) {Subpopulation 1:\\DE/current-to-best/1};
\node[subpop, below=of comm] (dyn2) {Subpopulation 2:\\DE/rand-to-best/2};
\node[subpop, below=of comm, xshift=3.5cm] (dyn3) {Subpopulation 3:\\DE/current-to-pbest/1};

\node[dashedbox, fit=(dyn1)(dyn2)(dyn3)] (dynbox) {};
\node[labeltext, right=0.2cm of dynbox.east] {Dynamic Operator Selection};

\draw[arrow] (init.south) -- (mutationbox.north);
\draw[arrow] (mutationbox.south) -- (diversitybox.north);
\draw[arrow] (diversitybox.south) -- (comm.north);
\draw[arrow] (comm.south) -- (dynbox.north);

\end{tikzpicture}
     
    \caption{The major components in proposed DE-MMOGC optimization algorithm}
    \label{fig:flow}   
\end{figure*}
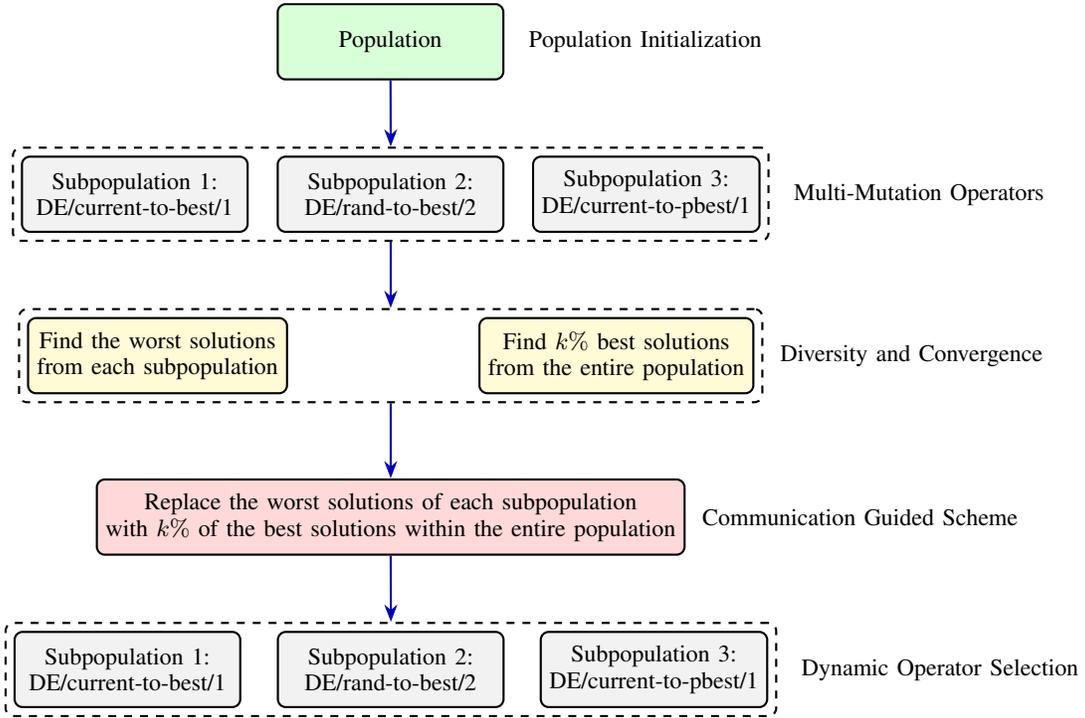

The proposed algorithm, DE-MMOGC, has four components: population initialization, multi-mutation operators, a communication guidance scheme, and dynamic operator selection. For better insight, Fig. \ref{fig:flow} illustrates the connection between the components. In the initial population, a total of $np$ solutions are generated randomly. The diversity and quality of the solutions are ensured in the later stages of the evolution by incorporating multiple operators of mutation \cite{Liu2020}. These mutation operators yield a target vector corresponding to the parent solution in the population. The initial population is split into three subpopulations, and each of them  undergoes one of the three DE mutation operators implemented here. The evolution of the sub-population takes place independently through the following operators: mutation, crossover, and selection, and eventually, the new solution is achieved. The proposed communication-guided scheme replaces the worst solution (with inferior fitness value) with the top solutions from each sub-population. Moreover, multi-mutation operators and the proposed communication-guided mechanism together improve the quality of solutions and maintain the diversity during the evolution process. The cycle is repeated until the maximum number of generations is reached.

\subsection{Population Initialization}

Initially, the potential $np$ candidate solutions are initialized randomly, which constitutes a population as
\begin{flalign}
      & x_{i} = x^{lower} + (x^{upper} - x^{lower}) \times rand & \\
       & i \in \{1,...,np \} &
\end{flalign}
where $x_{i}$ is a parent solution with a $rand$ random value between $[0,1]$. $x^{lower}$ and $x^{upper}$ represents lower space and upper space respectively.
Afterwards, the original population is divided into three subpopulations of varying dimensions, and one of the three DE mutation techniques discussed in the subsequent section is applied to each of them. 

\subsection{Multi-Mutation Operators of DE}

To generate a new mutant solution $v_{i}$, DE-MMOGC implements three standard mutation operators of DE. The applied mutation operators are as follows: 1) DE/current-to-best/1, 2) DE/rand-to-best/2, and 3) DE/current-to-$p$best/1. These operators are comprehensively discussed below.\\ 
``DE/current-to-best/1''
\begin{flalign}
    \label{1}
     & v_{i} = x_{i} + F (x_{best} - x_{i}) + F (x_{r1}-x_{r2}) &       \end{flalign}
``DE/rand-to-best/2''
\begin{flalign}
    \label{2}
     & v_{i} =  x_{r1}  + F (x_{best} - x_{r1}) + F (x_{r2} - x_{r3}) + F (x_{r1} - x_{r2}) &
\end{flalign}
``DE/current-to-$p$best/1''
\begin{flalign}
    \label{3}
    & v_{i} =  x_{i} + F (x_{pbest} - x_{i}) + F (x_{r1} - x_{r2}) &
\end{flalign}
The first mutation operator, ``DE/current-to-best/1'', pushes the solution set to the global best solution. Though it has most exploitation characteristics, it may result in premature convergence \cite{Wisittipanich2012}. On the other hand, the ``DE/rand-to-best/2'' is selected because of its maximum exploration capability but may have limited exploitation ability \cite{Zhao2016}. The last operator, known as ``DE/current-to-$p$best/1'', depends upon the archive of \textit{p\%} solutions evolved till previous generations. The current solution from the subpopulation is evolved from the best solution achieved so far. Hence, it has both exploitation and exploration characteristics \cite{Liu2020}. In this research, the top $30\%$ best solutions belonging to the archive of previous solutions are chosen. This arrangement of mutation strategies can lower the chances of local trapping, facilitating the equilibrium between exploration and exploitation of global search. $x_{r1},x_{r2}, x_{r3}$ are mutually exclusive and randomly selected solutions from the entire population. The new solution is obtained by the recombination of the mutant solution $v_{i}$ and target solution $x_{i}$ with a crossover rate $Cr$ 

\subsection{Communication Guided Scheme}

In contrast to single population-based algorithms that carry out a single mutation process, multi-mutation DE operates on three separate subpopulations that evolve independently of one another. These three subpopulations develop concurrently as long as the best fitness value so far is updated for a definite number of iterations. We propose a communication-guided scheme that shares the elite solutions across all subpopulations. After every generation, $k\%$ top solutions are selected from each subpopulation. This happens to replace the worst solution in the next generation by migrating the elite solution from each subpopulation. In this study, $10\%$ of the best solutions from each subpopulation are kept to guide the whole population towards the convergence. $x_{best}$ is randomly selected from that pool of best solutions. Additionally, it maintains the exploitation throughout the search process.       

\subsection{Dynamic Operator Selection}

We include dynamic operator selection based on the past success of the mutation operators. This enables the operators to improve their performance and takes the benefit to the persistent reuse in future generations. Consequently, the number of solutions assigned to each operator varies suitably after each generation. 
\par Suppose a mutation operator $s_i \in S$ is assigned for each subpopulation, where $S$ is the set of implemented mutation operators. After execution of an evolutionary process, update the probabilities for better-performing operators. We track the success rate as defined by the probability $p(s_i)$ if a trial solution $U_i$ replaces the individual solution $x_i$. The strategy adaptation takes place as
\begin{flalign}
    & p(s_i) = \frac{s_i}{\sum_j s_j} &
\end{flalign}

\section{DE-MMOGC for Improved CSM}
\label{demmogc}
In this section, we describe the real-world application of DE-MMOGC in optimizing CSM. Firstly, our simplified simulation model, WOFOST, is discussed, where two environmental stress factors along with other climatic inputs are considered that can resist the crop growth. Secondly, the data assimilation method, EnKF, is applied that can handle uncertainty and low-quality observation in a better way \cite{10504885}. Lastly, the process of fine-tuning weather parameters, $TAVG$ and $RAIN$, with DE-MMOGC, is described. We estimate LAI values with optimized weather parameters by both simulation model and ensemble-based assimilation method.       

\subsection{WOFOST Forecast}

For easy discussion, a simplified WOFOST model that predicts the LAI value for a crop is incorporated in this research. The modified WOFOST takes specific parameters for crop, weather, and soil data (listed in Table S1 of the Supplementary Material). Besides, it also accepts derived parameters that benefit the crop growth \cite{tia-0024-0032}. 

There are two derived input parameters in this simplified WOFOST: 1) temperature effect and 2) normalized water availability. The values of soil and crop parameters are assumed to be fixed, while weather parameters undergo evolution.    

The \textbf{temperature effect} above the base temperature is estimated for crop growth. It determines temperature-dependent potential by exceeding a threshold (TBASE) for heat that promotes crop growth. It is defined as the equation (\ref{temperature}),

    \begin{flalign}
    \label{temperature}
      & temperature\; effect =  TAVG - TBASE &
    \end{flalign}

The \textbf{normalized water availability} is computed, defined by the ratio of current available soil moisture to the maximum usable amount for the crop. The equation (\ref{water}) is given below,
\begin{flalign}
    \begin{split}
    \label{water}
      Soil\; moisture = (SMTAB\_MAX-WP)/(FC-WP) \\    
    water\; availability = \begin{bmatrix}
        RAIN & Soil\;moisture  
    \end{bmatrix}  
    \end{split}
\end{flalign}


Owing to these factors, we determine the time-step-based LAI growth, which is an essential part of CSMs. It directly depends on RGRLAI, IRRAD,  temperature effect, and water availability. A limiting factor is also introduced according to the LAI's current value about its maximum potential $(LAI\_MAX)$. 

\begin{ceqn}
    \begin{align}
    \label{deltalai}
    \begin{split}
      LAI = RGRLAI * temperature\;effect * water availability \\* IRRAD * (1 - current \; LAI/LAI\_MAX)  
    \end{split}
    \end{align}
\end{ceqn}

The equation (\ref{deltalai}) calculates the LAI growth for a time step, taking into account climate factors (temperature, water, and radiation), biological potential (RGRLAI), and dynamic growth stage (relative LAI). It simulates plant canopy development that is restrained logistically under various climate stresses.

\subsection{EnKF Assimilated LAI}

Suppose $a_t^i$ depicts assimilated LAI state variable of the $i$-th ensemble at time $t$ with noise $w_t^i$, and weather stress functions, $TAVG$ and $RAIN$, each ensemble member is propagated using the following equation (\ref{a}), 
\begin{flalign}
   \label{a}
     & a_{t|t-1}^i = f(a_{t-1},TAVG, RAIN) + w_t^i, \; w_t^i \sim \mathcal{N}(0,Q_t) &  
    \end{flalign}
Thereafter, various matrices (e.g., covariance and perturbations) are computed for Kalman Gain, and finally, assimilated LAI is determined. However, the procedure is being followed as described in Section S2-C (Ensemble Kalman Gain) in the Supplementary Material.

\subsection{Calibration with DE-MMOGC}
\label{sec:calibration}

Initialize a population $X$ of individuals representing weather parameters, $TAVG$, and $RAIN$ (we simply take $t_i$ for TAVG and $r_i$ for RAIN), 
    \begin{flalign}
       & X = \begin{bmatrix}
           t_1, t_2, t_3,...,t_{np} \\
           r_1,r_2,r_3,...,r_{np}
        \end{bmatrix}&
    \end{flalign}

The control parameters for the proposed DE-MMOGC include population size $np$ (usually the number of simulated days), generations $G$, mutation factor $F$, and crossover rate $Cr$. The given bounds for a candidate solution are denoted by the lower bound $L$ and upper bound $U$ of the respective parameter. Hence, for $i=1,2,..,np$
\begin{flalign}   
      &  t_i \in [L^{TAVG},U^{TAVG}] &\\
      &  r_i \in [L^{RAIN},U^{RAIN}] &
\end{flalign}    

\par The LAI prediction model $C(t+1)$ estimates the LAI values using parameters $X$. The true LAI for a specific crop is depicted by $y_i$. The goal of an objective function is to minimize the MSE between estimated LAI and actual LAI under parameter constraints. It is given by the equation (\ref{b}).
    \begin{flalign}
    \label{b}
       & min \; F (x) = \frac{1}{np} \sum_{i=1}^{np}(y_i - C(t+1)_i) ^2 &
    \end{flalign}
\par To summarize the above procedure for enhancing the CSM performance and estimating LAI by the proposed optimizer, Algorithm \ref{algo} is provided.


\begin{algorithm}
\begin{algorithmic}[1]
\caption{DE-MMOGC for assimilated LAI prediction }
\label{algo}

\FOR{\texttt{crop-variety}}

\STATE \textbf{Load Crop Data:}\\
\STATE Soil, Crop, and Weather Measurements (listed in Table S1 of the Supplementary Material)\\

\STATE \textbf{Initialize Simplified WOFOST:}\\
\STATE $LAI(0) \gets 0.01$\\
\STATE \textbf{Parameter Initialization for DE-MMOGC:} \\
\STATE np, G, F, CR, M, $[L^{TAVG},U^{TAVG}]$, $[L^{RAIN},U^{RAIN}]$ \\
\STATE $X = \begin{bmatrix}
    t_1,t_2,..,t_{np}\\
    r_1,r_2,...,r_{np}
\end{bmatrix}$ 
\STATE $t_i \in [L^{TAVG}, U^{TAVG}], r_i \in [L^{RAIN},U^{RAIN}]$

\FOR{\texttt{$generation \gets 1\; to\; G$}}
  \STATE Divide X into 3 subpopulations\\
  \STATE Assign defined mutation operators\\
   \STATE Apply DE-MMOGC as discussed in Section \ref{proposed}\\

   \STATE \textbf{Daily EnKF-assimilated LAI:}\\
   \FOR{\texttt{each day}}
        \FOR{\texttt{M ensemble member}}   
            \STATE \textbf{Daily Simulated LAI using WOFOST:} \\
            \STATE Temperature Effect using equation \ref{temperature}\\
            \STATE Water Availability using equation \ref{water}\\
            \STATE Estimate LAI using equation \ref{deltalai}\\
            \STATE Ensemble Propagation:\\
            $a_{t} = f(a_{t},TAVG,RAIN) + w_t, \; w_t \sim \mathcal{N}(0,Q_t) $\\
            \STATE Perturbed Observation:\\
            $y_t = \mathcal{H}_t a_t + v_t , \;  v_t \sim \mathcal{N}(0,R_t)$\\
            \STATE Kalman Gain:\\
            $K_t = P_{t}\mathcal{H}_t^T(\mathcal{H}_tP_{t}\mathcal{H}_t^T + R_t)^{-1}$\\
            \STATE Update Step:\\
             $a_{t+1} = a_{t} + K_t(y_t - \mathcal{H}(a_{t}))$\\
        \ENDFOR
   \STATE  $C(t+1)  = \frac{\sum_{i=1}^M C(\epsilon_i, t+1)}{M}$
  \ENDFOR
     \STATE \textbf{Evaluate MSE between assimilated LAI and observed LAI:}\\
    \STATE min $ MSE = \frac{1}{np} \sum_{i=1}^{np}(y_i - C(t+1)_i) ^2 $
\ENDFOR
  \STATE \textbf{Return:} \\
  \STATE MSE: Assimilated LAI and WOFOST Simulated LAI
\ENDFOR 
\end{algorithmic}
\end{algorithm}
\section{Result Analysis and Discussion}
\label{sec: discussion}

We evaluate the behavior of the DE-MMOGC optimizer across varieties of different crops, viz., wheat, rice, and cotton. We include $10\%$ perturbed observations, simulating the quality degradation in the data collection process—a common problem in agricultural practices. For clarity of presentation, the main findings are included in this section, whereas detailed experimental results and additional analyses are documented in the Supplementary Material (refer Section S4 (Result Analysis)) 
    
\subsection{Crop variety-wise analysis}

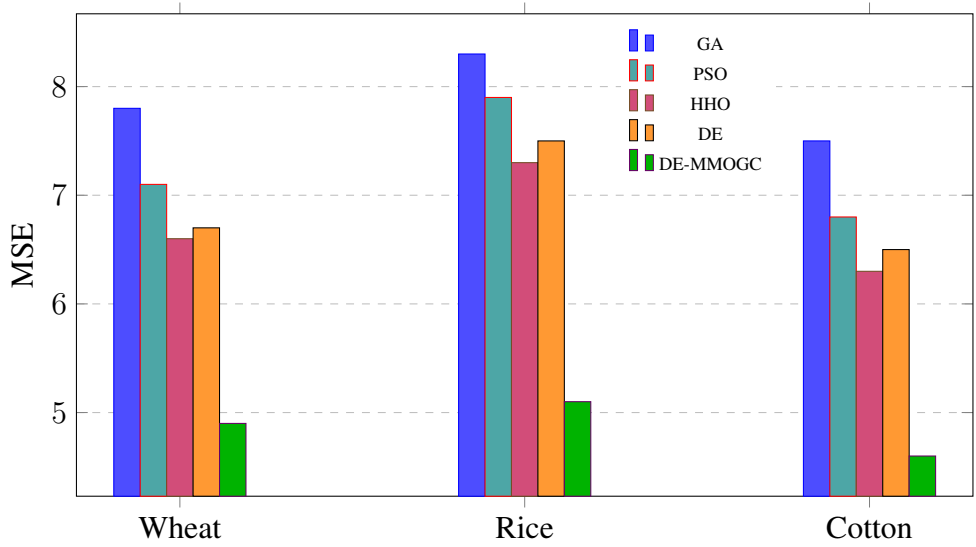
\begin{figure*}
\centering
\begin{tikzpicture}
\begin{axis}[
    ybar=0pt,
    bar width=10pt,
    width=13.5cm,
    height=8cm,
    enlarge x limits=0.15,
    ylabel={MSE},
    symbolic x coords={Wheat,Rice,Cotton},
    xtick=data,
    legend style={
        at={(0.78,0.98)},
        anchor=north east,
        font=\scriptsize,
        draw=none
    },
    ymajorgrids=true,
    grid style=dashed,
    tick label style={font=\large},
    label style={font=\large},
]

\addplot+[fill=blue!70] coordinates {
    (Wheat,7.8) (Rice,8.3) (Cotton,7.5)
};

\addplot+[fill=teal!70] coordinates {
    (Wheat,7.1) (Rice,7.9) (Cotton,6.8)
};

\addplot+[fill=purple!70] coordinates {
    (Wheat,6.6) (Rice,7.3) (Cotton,6.3)
};

\addplot+[fill=orange!80] coordinates {
    (Wheat,6.7) (Rice,7.5) (Cotton,6.5)
};

\addplot+[fill=green!70!black] coordinates {
    (Wheat,4.9) (Rice,5.1) (Cotton,4.6)
};

\legend{GA,PSO,HHO,DE,DE-MMOGC}

\end{axis}
\end{tikzpicture}
  \caption{Crop-wise MSE comparison of DE-MMOGC and other optimization algorithms. The figure illustrates the effectiveness of DE-MMOGC in minimizing prediction errors across different crop datasets and showing consistently lower MSE values.}
    \label{fig:graph2}
\end{figure*}

\begin{table*}[ht]
    \centering
    \caption{Performance metrics for the Wheat HD-2967 (Irrigated) variety}
    \label{tab:hd}
    \begin{tabular}{cccccc}
    \hline
    Algorithm &   &  MSE &  MAE & RMSE & Correlation\\
     \midrule 
     \multirow{2}{*}{DE}    &  Assimilation Metrics & 6.3115 & 1.9490 & 2.5123 & 0.9740\\
     & WOFOST Metrics & 6.4763 & 1.9746 & 2.5449 & 0.9711\\
     \hline

    \multirow{2}{*}{GA}    &  Assimilation Metrics & 7.8310 & 2.1535 & 2.7984 & 0.9792\\
     & WOFOST Metrics & 7.9476 & 2.1723 & 2.8191 & 0.9777\\ 
    \hline

     \multirow{2}{*}{HHO}    &  Assimilation Metrics & 8.3585 & 2.2621 & 2.8911 & 0.9698\\
     & WOFOST Metrics & 8.5294 & 2.2830 & 2.9205 & 0.9651\\ 
    \hline

    \multirow{2}{*}{PSO}    &  Assimilation Metrics & 6.6162 & 1.9915 & 2.5722 & 0.9758\\
     & WOFOST Metrics & 6.7249 & 2.0092 & 2.5932 & 0.9731\\ 
     \hline

     \multirow{2}{*}{DE-MMOGC}  &  Assimilation Metrics & \textbf{5.7587} & \textbf{1.8637} & \textbf{2.3997} & \textbf{0.9796}\\
     & WOFOST Metrics & 5.9484  & 1.8985 & 2.4389 & 0.9766\\
     
     \hline
    \end{tabular}  
\end{table*}

\begin{table*}[ht]
    \centering
    \caption{Performance metrics for the Wheat Lok-1 (Rainfed) variety}
    \label{tab:lok}
    \begin{tabular}{cccccc}
    \hline
    Algorithm &   &  MSE &  MAE & RMSE & Correlation\\
     \midrule
     \multirow{2}{*}{DE}    &  Assimilation Metrics & 4.7951 & 1.6949 & 2.1898 & 0.9786\\
     & WOFOST Metrics & 4.8984 & 1.7150 & 2.2132 & 0.9763\\
     \hline

    \multirow{2}{*}{GA}    &  Assimilation Metrics & 5.7270 & 1.8458 & 2.3931 & 0.9789\\
     & WOFOST Metrics & 5.8093  & 1.8618 & 2.4103 & 0.9772\\ 
    \hline
    
     \multirow{2}{*}{HHO}    &  Assimilation Metrics & 4.9345 & 1.7182 & 2.2214 & 0.9788 \\
     & WOFOST Metrics & 5.0461 & 1.7391 & 2.2464 & 0.9763\\ 
    \hline

    \multirow{2}{*}{PSO}    &  Assimilation Metrics & 4.9035 & 1.7157 & 2.2144 & 0.9749\\
     & WOFOST Metrics & 4.9789 & 1.7300 & 2.2314 & 0.9721 \\ 
     \hline

     \multirow{2}{*}{DE-MMOGC}  &  Assimilation Metrics &  \textbf{4.2758}  & \textbf{1.6071} & \textbf{2.0678} & \textbf{0.9792}\\
     & WOFOST Metrics & 4.4181 & 1.6377 & 2.1019 & 0.9759\\
     
     \hline
    \end{tabular}   
\end{table*}

Wheat is sensitive to temperature and water variations. For our study, we incorporate two varieties of wheat: HD-2967 (Irrigated) and Lok-1 (Rainfed). We applied DE-MMOGC and other algorithms for optimizing weather inputs to improve LAI prediction of both varieties. Tables \ref{tab:hd} and \ref{tab:lok} are presented for the performance metrics (MSE, MAE, RMSE, and correlation) for the respective varieties.  The detailed results for rice and cotton are presented in Section S4 (Result Analysis) of the Supplementary Material.

\par After observing these tables, we found DE-MMOGC outperformed the competitive algorithms. DE-MMOGC has a lower $30\%$ MSE compared to GA and $\sim 20\%$ enhancement over single mutation DE. This successful behavior showed that communication-guided scheme across subpopulations support escaping local optima commonly met in earlier optimization algorithms.  

\par The bar chart provided in Fig. \ref{fig:graph2} shows the crop-wise MSE analysis among all optimization algorithms. We observe that DE-MMOGC outperforms all algorithms. It shows promising results for each crop variety—wheat, rice, and cotton. The most promising results are recorded in BT Cotton (RCH 134), where DE-MMOGC has the lowest predictive error. Even in other scenarios (wheat and rice), DE-MMOGC continues to have a clear edge, suggesting reliable and steady optimization in a range of farming methods.

\par To summarize the crop-wise analysis, we found DE-MMOGC remains consistent across all crop varieties. It demonstrates the improvements in error metrics for LAI-based predictive models compared to GA and single-mutation evolutionary algorithms. DE-MMOGC is ideal for real-world crop modelling applications where data can be high-dimension, noisy, and incomplete.      

\subsection{Assimilation and No Assimilation }

\begin{figure*}
    \centering
    \subcaptionbox{Wheat HD-2967 (Irrigated)}{\includegraphics[width=\linewidth,trim = 0cm 23cm 1cm 1cm,clip]{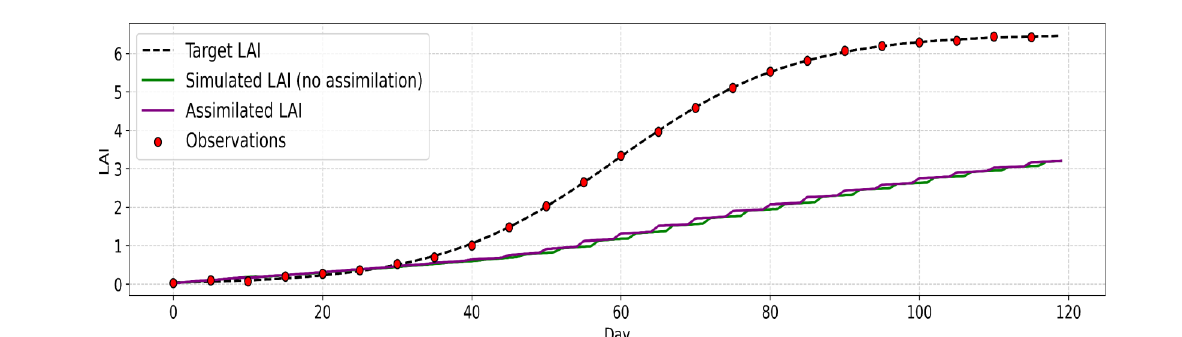}}
    \subcaptionbox{Wheat Lok-1 (Rainfed)}{\includegraphics[width=\linewidth,trim = 0cm 24cm 1cm 1cm,clip]{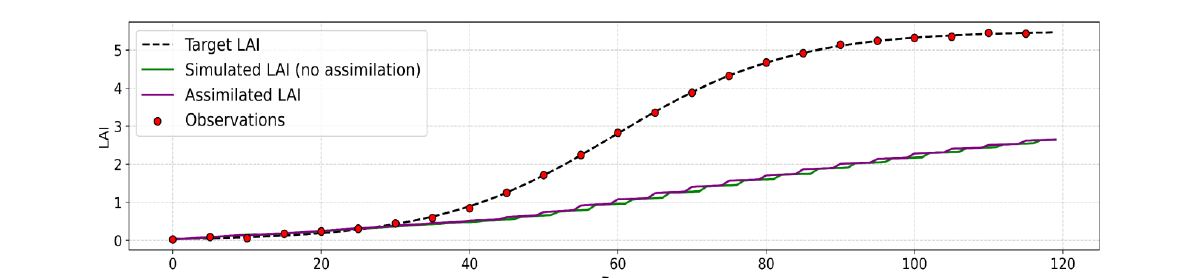}}
    
    \caption{Convergence graphs generated by DE-MMOGC for improving LAI estimation of wheat crops. The graphs specifically illustrates the LAI trajectories for 120 simulated days}
    \label{fig:convergence1}
\end{figure*}

In order to assess the value added by simulating real-time observation data, in terms of noise or uncertainty, we compare DE-MMOGC's performance on two fronts:
\begin{enumerate}
    \item \textbf{WOFOST Forecast}: Predict LAI using simplified WOFOST simulation model without any missing observations or noisy data

    \item \textbf{EnKF Assimilation}: Estimate LAI with the integration of EnKF to assimilate real-time observations
\end{enumerate}

By analyzing the results recorded in tables for all crop varieties, we found that assimilation improves the LAI-based yield estimation. It showcases its efficacy and robustness in uncertain and noisy environments. Moreover, the integration of EnKF results in stable and consistent LAI prediction during dynamic growth stages when the simulation model, WOFOST, tends to diverge. Coming to other metrics, like the correlation coefficient, we have seen a significant increase in the metric with assimilation, proving improved alignment with the real field conditions. EnKF uses ensemble propagation to fill in missing or noisy LAI observations (caused by cloud cover or sensor oversight, for example), preserving prediction quality even when 10\% of the data is missing.   

\par The comparison between WOFOST simulation and EnKF assimilation is made to simulate the real-time environmental adaptations. As opposed to WOFOST crop simulation, DE-MMOGC, with EnKF assimilation, enhances the predictive accuracy and reliability for real-time agricultural decision support.


\subsection{Scalability and Convergence}

With no discernible decline in convergence rate, the DE-MMOGC method exhibits scalable performance across a range of simulation durations and population sizes. In contrast to baselines, the communication and diversity preservation strategies produce faster and smoother convergence curves by preventing stagnation, particularly during extended optimization runs. 

\par The performance of the implemented DE-MMOGC algorithm in optimizing weather parameters (temperature and rainfall) for crop simulation under wheat farming scenarios is illustrated by the convergence graph shown in Fig. \ref{fig:convergence1}. The simulated LAI without assimilation understates the actual crop canopy development throughout the course of the growing season for the wheat HD-2967 variety. The overall trend is steady, but after day 50, it starts to diverge considerably. However, especially between days 40 and 100, the assimilated LAI curve closely matches the goal LAI and the observed data, demonstrating how well the EnKF integration corrects model bias brought on by noisy or suboptimal weather conditions. The EnKF-assimilated LAI trajectory for the rainfed variety exhibits a substantially lower error margin, particularly between days 40 and 100, even if it is still marginally below the target LAI. This suggests that the model was able to adjust to water-limited scenarios since DE-MMOGC was able to optimize certain parameters. To maintain the predictions close to the actual LAI values, the EnKF filter offers dynamic adjustment. The significance of data assimilation in unpredictable and dynamic situations is highlighted by the rainfed scenario. The enhanced LAI convergence makes it clear that the suggested strategy is not only successful in stable (irrigated) conditions but also resilient in stressful and stochastic weather scenarios. 

\par The convergence plots corresponding to rice and cotton are illustrated in the Section S4-A (Convergence Plots for Rice and Cotton) of the Supplementary Material.

\section{Conclusion and Future Directions}
\label{sec: conclusion}

Single-mutation DE algorithms have limited adaptability due to their reliance on a fixed mutation approach, which frequently results in premature convergence and decreased diversity in different optimization landscapes. To overcome these drawbacks, this research proposes DE-MMOGC, a new multi-mutation DE variant combined with a communication-guided scheme and dynamic operator selection, tailored for optimizing parameters in uncertain spaces.
\par Experiments show that DE-MMOGC significantly performs better than traditional single-mutation DE variants. The proposed approach specifically has 30\% lower MSE compared to GA and $\sim 20\%$ enhancement over single-mutation DE.  In a number of benchmark situations, DE-MMOGC demonstrated the efficacy of communication-guided operator cooperation by maintaining better solution quality even in the face of uncertain conditions.
\par While this research has demonstrated notable enhancements in optimizing CSM under diverse weather conditions, several impactful avenues remain open for exploration. Future studies can work with comprehensive CSMs such as DSSAT or APSIM which simulate a wider variety of physiological functions and agricultural management. In real farming scenarios, it involves multi-objective optimization, balancing conflicting objectives, such as maximum grain production and greenhouse gas minimization. Extending DE-MMOGC to a multi-objective framework could strengthen the analysis for sustainable agricultural practices.

\end{document}


\maketitle

\section{Crop Simulation Models}
CSMs are complex computational tools that simulate how crops grow, develop, and yield in response to management techniques (planting date, irrigation, fertilizer), and environmental factors such as weather, and soil. There are common CSMs in agricultural research including DSSAT \cite{dssat}, APSIM \cite{apsim}, and WOFOST \cite{WOFOST}. DSSAT has large library of 40 crop models. It is well-structured and beneficial for management optimization and field-based research. On the other hand, APSIM emphasis on soil dynamics, fallows, intercrops, and the capacity to model intricate cycles. It can be used for farming practices research and formulating climate adaptive strategies. Considering WOFOST, it is highly suitable for regional yield forecasting and crop monitoring. 

\par Despite their strengths, these CSMs have certain limitations that researchers should carefully consider. These models are extremely complex with several parameters. This makes them difficult to understand and fine-tune. Simulating environmental stress factors such as heatwaves and drought remains a major challenge. Model predictive accuracy can be reduced on uncertain weather inputs and climate projections.

\section{Background}
\label{prelim}

The basic concepts of multi-mutation DE optimization, simplified WOFOST for crop simulation, and the EnKF assimilation method are explained in this section. 

\subsection{Multi-mutation DE Optimization}
\label{multi-mutation}
Starting with a population of uniformly randomly distributed candidate solutions drawn from the search space, each solution develops over time through iterations known as \textit{generations}. Crossover, selection, and mutation are the sub-processes by which evolution occurs. These sub-processes are repeated until certain conditions are met \cite{sallam2018,sallam2020}.


Several mutation operators have been proposed in literature that work cohesively in order to balance exploration-exploitation equilibrium in DE. In general, the following mutation operators are employed \cite{Liu2020}.
\\``DE/rand/1''

\begin{flalign}
   &V_{np}^G = X_{r1}^G + F(X_{r2}^G-X_{r3}^G)&
\end{flalign}
``DE/current/1''
\begin{flalign}
        &V_{np}^G = X_{np}^G + F(X_{r1}^G-X_{r2}^G)& 
\end{flalign}
``DE/best/1''
\begin{flalign}
     &V_{np}^G = X_{best}^G + F(X_{r1}^G-X_{r2}^G)&    
\end{flalign}
``DE/rand/2''
\begin{flalign}
        &V_{np}^G = X_{r1}^G + F(X_{r2}^G-X_{r3}^G) + F(X_{r4}^G-X_{r5}^G)& 
\end{flalign}
``DE/best/2''
\begin{flalign}
     &V_{np}^G = X_{best}^G + F(X_{r1}^G-X_{r2}^G) + F(X_{r3}^G-X_{r4}^G)&   
    \end{flalign}
``DE/current-to-best/1''
\begin{flalign}
         &V_{np}^G = X_{np}^G + F(X_{best}^G-X_{np}^G) + F(X_{r1}^G-X_{r2}^G)&  
    \end{flalign}
``DE/current-to-rand/1''
\begin{flalign}
         &V_n^G = X_{n}^G + F(X_{r1}^G-X_{n}^G) + F(X_{r2}^G-X_{r3}^G)&  
\end{flalign}
``DE/current-to-$p$best/1''
\begin{flalign}
         &V_{np}^G = X_{np}^G + F(X_{pbest}^G-X_{np}^G) + F(X_{r1}^G-X_{r2}^G)&  
\end{flalign}
The selection criteria for choosing parent solutions (${r1}, {r2}, {r3}, {r4}, {r5}$) from the population $X$ vary in the literature \cite{xmode}. $X_{best}^G$ is the \textit{best} solution set selected from the entire population $X$ at generation $G$. $X_{pbest}^G$ is the \textit{best} solution set taken from the archive of $p\%$ solutions evolved till the last generations. $F$ is a scaling factor that ensures the population diversity. By carefully selecting mutation mechanisms and setting the criteria for parent solutions, one can prevent becoming trapped in local optima. To generate the offspring, the set of mutant solutions $V_{np}^G$ and target solutions $X_{np}^G$ at generation $G$ are recombined with a crossover rate $Cr$, and a trial solution $U_{np}^G$ is obtained.

\subsection{Simplified WOFOST Simulation Model }
The annual growth and yield of field crops can be quantitatively analyzed using the WOFOST (WOrld FOod STudies) \cite{WOFOST} simulation model. In order to restrict and modify the crop-growing process using soil and crop parameter data, the model is powered by daily meteorological data. In this research, the authors offer a simplified version of the WOFOST model that can simulate crop growth under weather effects. 

LAI, which is the multiple of the total area of plant leaves on the land per unit of land area, is a critical indicator of the crop's potential grain yield because it shows how well the crop absorbs light and assimilates carbon. Thus, our simplified WOFOST model predicts the LAIs of crops. To predict LAI at time $(t + \Delta t)$, simplified WOFOST accepts two types of input: first is the LAI value at time $t$, and second are crop measurements, soil parameters, and weather conditions that can manipulate the crop-growing process. The specific parameters for weather, crop, and soil that are considered in this research (not limited to) are listed in Table \ref{tab:wofostpar}. For instance, the specific measurements for crop parameters include base temperature, relative growth rate of LAI, maximum LAI, and moisture content.

\begin{table}[]
    \centering
    \caption{Different parameter measurements for LAI prediction by simplified WOFOST}
    \label{tab:wofostpar}
    \begin{tabular}{p{0.15\linewidth} p{0.20\linewidth} p{0.35\linewidth} p{0.15\linewidth} }
    \hline
     Parameter Name  &  Specific Measurement &  Description &  Unit \\
     \hline
    \multirow{3}{*}{Soil}  &  FC & Field capacity shows soil water content & pF\\
    & WP & Wilting Point & / \\
    & BD & Bulk Density & / \\ 
      \hline

      \multirow{6}{*}{Crop} & Crop Name & Name of the crop for growth prediction & /\\
       & Variety & Variety of the crop & / \\
       & TBASE & Base temperature for crop growth & \textdegree C \\
       & RGRLAI & Maximum relative growth rate of LAI & ha/ha.d \\
       & LAI\_MAX & Maximum LAI for virtual crop & / \\
       & SMTAB\_MAX & Initial maximum moisture content in rooting & pF\\
       \hline

      \multirow{3}{*}{Weather} & TAVG & Daily average temperature & \textdegree C \\
      & IRRAD & Daily radiation & J/$m^2$/day \\
      & RAIN & Precipitation & mm \\
     \hline
    \end{tabular}
    
\end{table}

\subsection{Ensemble Kalman Filter}
\label{EnKF}

Developed for high-dimensional states and nonlinear systems (such as weather and crop simulations), the EnKF is a Monte Carlo approximation \cite{10.1145/2379810.2379816} of the Kalman Filter \cite{10.1145/3519935.3520050, 10.1145/3363294}. Covariance matrices are estimated using an ensemble of state vectors. A linear dynamic system state is estimated from noisy data using the recursive EnKF technique. There are two important functions in EnKF: \textit{predict} and \textit{update}. For a simple discussion, a numerical example is given in Appendix \ref{appendix}, and the steps of the algorithm are thoroughly explained below:

\subsubsection{Ensemble Representation}
Initialize \textit{M} ensemble members $\{{a_t}^i\}_{i=1}^M$ from a normal distribution $\mathcal{N}(\mu,P_t)$ where $\mu$ is the mean and $P_t$ is the covariance matrix. Here, $\{a_t\}$ represents the state variable at time $t$.

\subsubsection{Prediction}
In EnKF, we add process noise $w_t^i$ to the state observations. After initializing the state variable and process noise, each ensemble member is further propagated, and subsequently, the mean and covariance are updated. The propagation is performed as follows:  
      \begin{flalign}
            \label{ensemble}
                &a_{t}^i = \mathcal{M}_t(a_{t}^i) + w_t^i, w_t^i \sim \mathcal{N}(0,Q_t)&
        \end{flalign}
      where, $\mathcal{M}_t$ is a non-linear state transition function. $w_t^i$ is a process noise with covariance $Q_t$ that models uncertainty on cloudy or dusky days.

Compute the mean $\mu$ and covariance matrix $P_t$, which will be used by the Kalman filter. 
\begin{flalign}
    &\mu_{t}    = \frac{1}{M}\sum_{i=1}^M a_{t}^i&
\end{flalign}         
\begin{flalign}
  &P_{t} = \frac{1}{M-1}\sum_{i=1}^M (a_{t}^i - \mu_{t})(a_{t}^i-\mu_{t})^T &  
\end{flalign}                           
\subsubsection{Update} 
Then, we have state observation of LAI at time $t$, $y_t^i$. Like process noise, an observation noise $v_t^i$ with covariance $R_t$ and mean $0$ is added to create perturbed observations as follows:
  \begin{flalign}
      &y_t^i = \mathcal{H}_t a_t^i + v_t^i , \;  v_t^i \sim \mathcal{N}(0,R_t)& 
  \end{flalign}           
$\mathcal{H}_t$ is the linear operator that signifies the number of observed values that are currently not involved in assimilation. It is employed to illustrate how the simulated state and the observed quantity are linearly related \cite{10.1145/3519935.3520050}. 

By the covariance vectors $P_t, R_t$ and linear operator $H_t$, now calculate the ensemble version of Kalman Gain $K_t$ as follows.
\begin{flalign}
 &K_t = P_{t|t-1}\mathcal{H}_t^T(\mathcal{H}_tP_{t|t-1}\mathcal{H}_t^T + R_t)^{-1}&   
\end{flalign}             
Here, $\mathcal{H}_t^T$ is a transpose of $\mathcal{H}$. 

Finally, the results of EnKF assimilation can be computed as follows:
\begin{flalign}
   \label{update_step}
               &a_{t+1}^i = a_{t}^i + K_t(y_t^i - \mathcal{H}(a_{t}^i))&  
\end{flalign}
Now, $a_t^i$ denotes the predicted LAI result for time $t$ by EnKF, $K_t$ is a Kalman Gain, and $y_t^i$ is a perturbed observation. 

\section{Experimental Setup}
DE-MMOGC is applied to improve the crop development model by adjusting weather parameters. The setup is run in two scenarios: \textit{with} and \textit{without} EnKF assimilation. It has been demonstrated that the EnKF assimilation predictive model outperforms the streamlined WOFOST simulation. For comparative analysis, four existing evolutionary optimization algorithms are selected, namely 1) Differential Evolution \cite{Storn1997}, 2) Genetic Algorithm (GA) \cite{632992}, 3) Particle Swarm Optimization (PSO) \cite{488968}, and 4) Harris Hawk Optimization (HHO) \cite{HEIDARI2019849}. Furthermore, the details of experimental data and other settings are explained in successive sub-sections.
\subsection{Experimental Data}
The accumulation of data relevant to crops and their varieties is done from the repository of the Indian Council of Agricultural Research (ICAR) \cite{icar}. It lists climatic conditions for different varieties of a crop \cite{icar2022rabi}. Let's say the wheat crop has two different varieties, viz., HD-2967 (Irrigated) and Lok-1 (Rainfed). Their weather conditions and phenological features are completely different from each other. For better insight, some of the crop, soil, and weather measurements pertaining to wheat, rice, and cotton varieties are provided in Table \ref{tab:wheat}. 
However, the proposed optimizer is assessed on three crops, i.e., wheat, rice, and cotton, to build up the predictive accuracy of simplified WOFOST after getting optimized temperature and rainfall parameters.
\begin{table*}[]
    \centering
    \caption{Crop varieties and their specific measurements}
    \label{tab:wheat}
    \begin{tabular}{p{0.08\linewidth} p{0.15\linewidth} p{0.20\linewidth} p{0.20\linewidth} p{0.20\linewidth}}
    \hline
      Crop   &  Variety & Crop Parameters & Soil Parameters & Weather Parameters\\
     \hline    
      \multirow{2}{*}{Wheat}   & HD-2967 (Irrigated) & LAI\_MAX:6.5 RGRLAI:0.035 TBASE :0 \textdegree C & FC:0.30 WP:0.12 BD:1.3 & TAVG: 20-25\textdegree C Rainfall: 400-600 mm\\
      \cline{2-5}\\
      
       & Lok-1 (Rainfed) & LAI\_MAX:5.5 RGRLAI:0.028 TBASE: 10\textdegree C & FC:0.28 WP:0.11 BD:1.4 &  TAVG: 18-22 \textdegree C Rainfall: 300-500 mm\\
      \hline

      \multirow{2}{*}{Rice} & IR64 (Lowland) & LAI\_MAX:7.0 RGRLAI:0.04 TBASE:10 \textdegree C & FC:0.34 WP:0.15 BD:1.2 & TAVG:25-30 \textdegree C Rainfall: 800-1200 mm\\
      \cline{2-5}\\

      & Sahbhagi Dhan (Upland) & LAI\_MAX:6.0 RGRLAI:0.035 TBASE:10 \textdegree C & FC:0.29 WP:0.13 BD:1.3 & TAVG:24-28 \textdegree C Rainfall: 500-900 mm\\
      
     \hline

     Cotton & BT Cotton (RCH 134) & LAI\_MAX:5.5 RGRLAI:0.03 TBASE:15 \textdegree C & FC:0.28 WP:0.12 BD:1.4 & TAVG:25-35 \textdegree C Rainfall: 500-900 mm \\
      \hline
    \end{tabular}
    
\end{table*}

\subsection{Parameter Settings}
 
The variables for EnKF and other common parameters for both DE-MMOGC and alternative algorithms are listed in Tables \ref{tab:enkf} and \ref{tab:parameter} respectively. For EnKF, the  true LAI value is observed after every five days. Moreover, we incorporate $10\%$ noisy data that can simulate low-quality observations or missing observations during cloudy weather or heavy rainfall.  

\begin{table}[]
    \centering
    \caption{EnKF variable settings in CSM}
    \label{tab:enkf}
    \begin{tabular}{p{0.10\linewidth} p{0.30\linewidth} p{0.35\linewidth}}
     \hline    
     Variable   &  Description  &   Initial Value\\
    \hline
      $a_t$   &  State Variable: LAI & 0.01 \\
      $y_t$ & Observed LAI & specific to variety \\
      $d$ & Simulated days & 120\\
      $M$ & Ensemble Size & 50\\
      $R_t$ & Observation noise & 0.1 \\
      $H_t$ & Maps model state to observations & 1\\
    \hline  
    \end{tabular}  
\end{table}

\begin{table}[]
    \centering
    \caption{Parameter settings for DE-MMOGC and other comparative optimization algorithms}
    \label{tab:parameter}
    \begin{tabular}{p{0.15\linewidth} p{0.10\linewidth} p{0.30\linewidth} p{0.15\linewidth}}
     \hline    
     Algorithm & Parameter   &  Description  &   Initial Value\\
    \hline

      \multirow{5}{*}{DE} & np & population size & 10\\
         & G & maximum iterations & 50\\
         & strategy & best/1/bin & -- \\ 
         & F & scaling factor & [0.5,1]\\
         & CR & crossover rate & 0.7 \\
      \hline

      \multirow{6}{*}{GA} & np & population size & 10\\
         & G & generations & 50\\
         & strategy & Gaussian mutation & $\mu =0$, $\sigma = 1$  \\
         & selection & Tournament  & --\\ 
         & $\alpha$ & blending & 0.5\\
         & indpb & individual probability & 0.2\\

       \hline
     
     \multirow{2}{*}{PSO} & np & swarm size & 10\\
         & G & maximum iterations & 50\\
      \hline
      
     \multirow{3}{*}{HHO} & np & number of hawks & 10\\
         & G & maximum iterations & 50\\
         & r,q & soft and hard besiege parameters & [0,1]\\
     \hline
     
        \multirow{5}{*}{\textbf{Proposed}} & np & population size & 10\\
         & G & maximum iterations & 50\\
         & $p$ & $p$ in current-to-pbest mutation operator & 30\%\\
         & $x_{best}$ & best solutions & 10\%\\
         & F, Cr & scaling and crossover & 0.6, 0.9\\
     
    \hline  
    \end{tabular}  
\end{table}

\subsection{Evaluation Metrics}

Crop simulation models are assessed for their prediction accuracy in representing observed data using a range of statistical indicators. In our research, we incorporate the following metrics: MSE, mean absolute error (MAE), root mean squared error (RMSE), and correlation.

\section{Result Analysis}

\par The data collection for rice, another crop for simulation in this research, usually gets affected by rainfall observations. It can have an adverse impact on the performance of baseline algorithms. Considering rice varieties, for IR64 (Lowland) and Sahbhagi Dhan (Upland), experimental results in Tables \ref{tab:ir64} and \ref{tab:upland} are provided. It shows DE-MMOGC remains stable with perturbations, owing to the integration with EnKF. Existing algorithms (PSO and HHO) display poor convergence, while DE-MMOGC produces a faster and more consistent convergence by maintaining diversity and dynamically modifying mutation operators.
\par Considering cotton crop simulation, it is highly sensitive to temperature and soil moisture. For the cotton crop, the BT Cotton (RCH 134) variety is considered, for which performance metrics are recorded in Table \ref{tab:cotton}. As observed in the table, the proposed optimizer DE-MMOGC achieves better MSE and high correlation (Correlation $>0.97$) compared to standard DE and existing optimizers. The multi-mutation plays a significant role in preserving the diversity and improving the exploration property. Along with that, a communication-guided scheme helps in enhancing exploitation characteristics without compromising convergence stability.

\begin{table*}[h]
    \centering
    \caption{Performance metrics for the Rice IR64 (Lowland) variety}
    \label{tab:ir64}
    \begin{tabular}{cccccc}
    \hline
    Algorithm &   &  MSE &  MAE & RMSE & Correlation\\
     \hline 
     \multirow{2}{*}{DE}    &  Assimilation Metrics & 7.6593 & 2.1361 & 2.7675 & 0.9794\\
     & WOFOST Metrics & 7.8038 & 2.1591 & 2.7935 & 0.9795\\
     \hline

    \multirow{2}{*}{GA}    &  Assimilation Metrics & 8.9816 & 2.3109 & 2.9969 & 0.9775\\
     & WOFOST Metrics &  9.1175 & 2.3318 & 3.0195 & 0.9757\\ 
    \hline

     \multirow{2}{*}{HHO}    &  Assimilation Metrics & 8.9610 & 2.3500 & 2.9935 & 0.9760\\
     & WOFOST Metrics & 9.1722 & 2.3765 & 3.0286 & 0.9711\\ 
    \hline

    \multirow{2}{*}{PSO}    &  Assimilation Metrics & 7.6998 & 2.1487 & 2.7748 & 0.9757\\
     & WOFOST Metrics & 7.8307 & 2.1684 & 2.7983 & 0.9729\\ 
     \hline

     \multirow{2}{*}{DE-MMOGC}  &  Assimilation Metrics & \textbf{6.6841} & \textbf{2.0085} & \textbf{2.5854} & \textbf{0.9814}\\
     & WOFOST Metrics & 6.9054 & 2.0460 & 2.6278 & 0.9763 \\
     
     \hline
    \end{tabular}    
\end{table*}

\begin{table*}[h]
    \centering
    \caption{Performance metrics for the Rice Sahbhagi Dhan (Upland) variety}
    \label{tab:upland}
    \begin{tabular}{cccccc}
    \hline
    Algorithm &   &  MSE &  MAE & RMSE & Correlation\\
     \hline 
     \multirow{2}{*}{DE}    &  Assimilation Metrics & 5.0886 & 1.7571 & 2.2558 & 0.9754\\
     & WOFOST Metrics & 5.2355 & 1.7839 & 2.2881 & 0.9717\\
     \hline

    \multirow{2}{*}{GA}    &  Assimilation Metrics & 6.6215 & 1.9876 & 2.5732 & 0.9790\\
     & WOFOST Metrics & 6.7204 & 2.0055 & 2.5924 & 0.9795\\ 
    \hline

     \multirow{2}{*}{HHO}    &  Assimilation Metrics & 6.8842 & 2.0917 & 2.6238 & 0.9595\\
     & WOFOST Metrics & 7.0953 & 2.1258 & 2.6637 & 0.9538\\ 
    \hline

    \multirow{2}{*}{PSO}    &  Assimilation Metrics & 5.7383 & 1.8560 & 2.3955 & 0.9753\\
     & WOFOST Metrics & 5.8306 & 1.8722 & 2.4147 & 0.9726\\ 
     \hline

     \multirow{2}{*}{DE-MMOGC}  &  Assimilation Metrics & \textbf{5.0706} & \textbf{1.7499} & \textbf{2.2518} & \textbf{0.9811}\\
     & WOFOST Metrics & 5.2379 & 1.7828 & 2.2886 & 0.9758\\
     
     \hline
    \end{tabular}   
\end{table*}  
\begin{table*}[h]
    \centering
    \caption{Performance metrics for the BT Cotton (RCH 134) variety}
    \label{tab:cotton}
    \begin{tabular}{cccccc}
    \hline
    Algorithm &   &  MSE &  MAE & RMSE & Correlation\\
     \hline 
     \multirow{2}{*}{DE}    &  Assimilation Metrics & 5.2148 & 1.7701 & 2.2836 & 0.9768\\
     & WOFOST Metrics & 5.3022 & 1.7864 & 2.3027 & 0.9741\\
     \hline

    \multirow{2}{*}{GA}    &  Assimilation Metrics & 5.8086 & 1.8615 & 2.4101 & 0.9785 \\
     & WOFOST Metrics & 5.8930 & 1.8774 & 2.4275 & 0.9801\\ 
    \hline

     \multirow{2}{*}{HHO}    &  Assimilation Metrics & 5.6994  & 1.8435 & 2.3873 & 0.9783\\
     & WOFOST Metrics & 5.7867 & 1.8585 & 2.4056 & 0.9758\\ 
    \hline

    \multirow{2}{*}{PSO}    &  Assimilation Metrics & 5.1320 & 1.7584 & 2.2654 & 0.9733\\
     & WOFOST Metrics & 5.2088 & 1.7727 & 2.2823 & 0.9704\\ 
     \hline

     \multirow{2}{*}{DE-MMOGC}  &  Assimilation Metrics & \textbf{4.4139}  & \textbf{1.6358} & \textbf{2.1009} & \textbf{0.9817}\\
     & WOFOST Metrics & 4.5635 & 1.6675 & 2.1362 & 0.9749\\
     
    \hline
    \end{tabular}   
\end{table*}

\subsection{Convergence Plots for Rice and Cotton }
\begin{figure*}[h]
    \centering
    \subcaptionbox{Rice IR64 (Lowland)}{\includegraphics[width=\linewidth,trim = 0cm 24cm 1cm 1cm,clip]{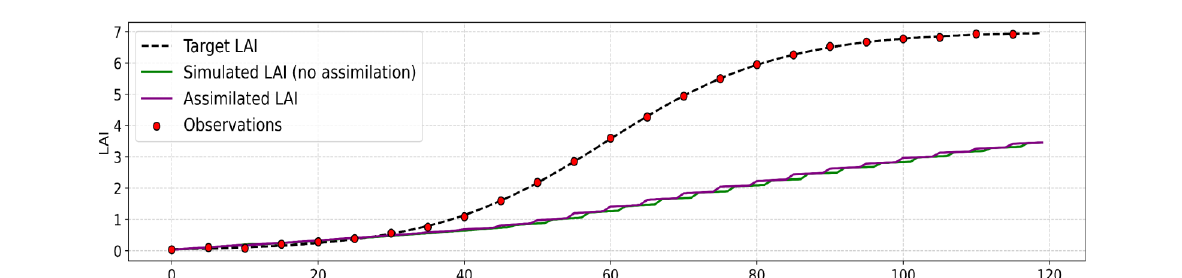}}
    \subcaptionbox{Sahbhagi Dhan (Upland)}{\includegraphics[width=\linewidth,trim = 0cm 24cm 1cm 1cm,clip]{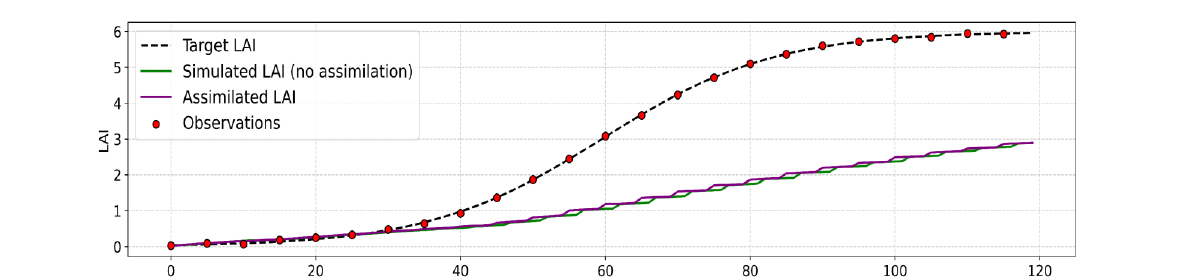}}
    
    \caption{Convergence graphs generated by DE-MMOGC for improving LAI estimation of rice varieties. The graphs specifically illustrates the LAI trajectories for 120 simulated days}
    \label{fig:convergence3}
\end{figure*}

\begin{figure*}[h]
    \centering
    \includegraphics[width=\linewidth,trim = 0cm 24cm 1cm 1cm,clip]{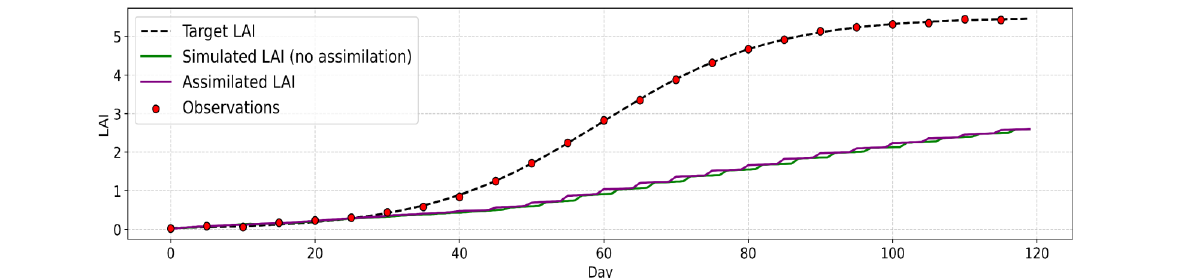}
    \caption{Convergence graphs generated by DE-MMOGC for improving LAI estimation of cotton variety. The plot specifically illustrates the LAI trajectories for 120 simulated days}
    \label{fig:convergence2}
\end{figure*}

The convergence plots in Fig. \ref{fig:convergence3} show the assimilation and optimization results for two opposite rice ecosystems: Sahbhagi Dhan (Upland) and IR64 (Lowland). Talking about the former variety, it is grown for drought-resistant situations. In the convergence graph, the simulated LAI (represented by the orange line) is below the target LAI, though structurally consistent. The refined plot produced by DE-MMOGC and EnKF (represented by the green line) depicts the significant improvement and target LAI, especially between $40-90$ simulated days. This reveals the strengths of DE-MMOGC and EnKF in boosting the accuracy of crop state representation in rainfed upland and stress-prone conditions. Considering the latter rice variety, suitable for irrigated lowland environments, demonstrates more active crop growth. The rainfall parameter, optimized by DE-MMOGC, indicates substantial variability, which the EnKF method reconciled by assimilating real values. In brief, the hybrid DE-MMOGC and EnKF framework shows the performance boost in both rice varieties under stress conditions.        

\par Fig. \ref{fig:convergence2} presents the convergence behavior of the proposed DE-MMOGC on the LAI prediction over the BT Cotton RCH 134 growth cycle. As observed in the convergence plot, there is a noticeable improvement in the EnKF-assimilated LAI (green line), compared to the baseline simulation. In order to dynamically correct the trajectory of the simulated state, the data assimilation mechanism effectively incorporates real-time observational data (red dots). The target LAI trajectory's alignment with the red observation markers validates the predictive accuracy of the reference data. A low residual error against these observed points is seen in the assimilated LAI curve, demonstrating successful convergence and efficient application of field-based assimilation.

\subsection{Comparative Analysis with Traditional Optimizers}

\begin{figure*}[t]
    \centering
    \subcaptionbox{MSE}[0.48\linewidth]{\includegraphics[width=\linewidth,trim=0cm 16cm 2cm 1cm,clip]{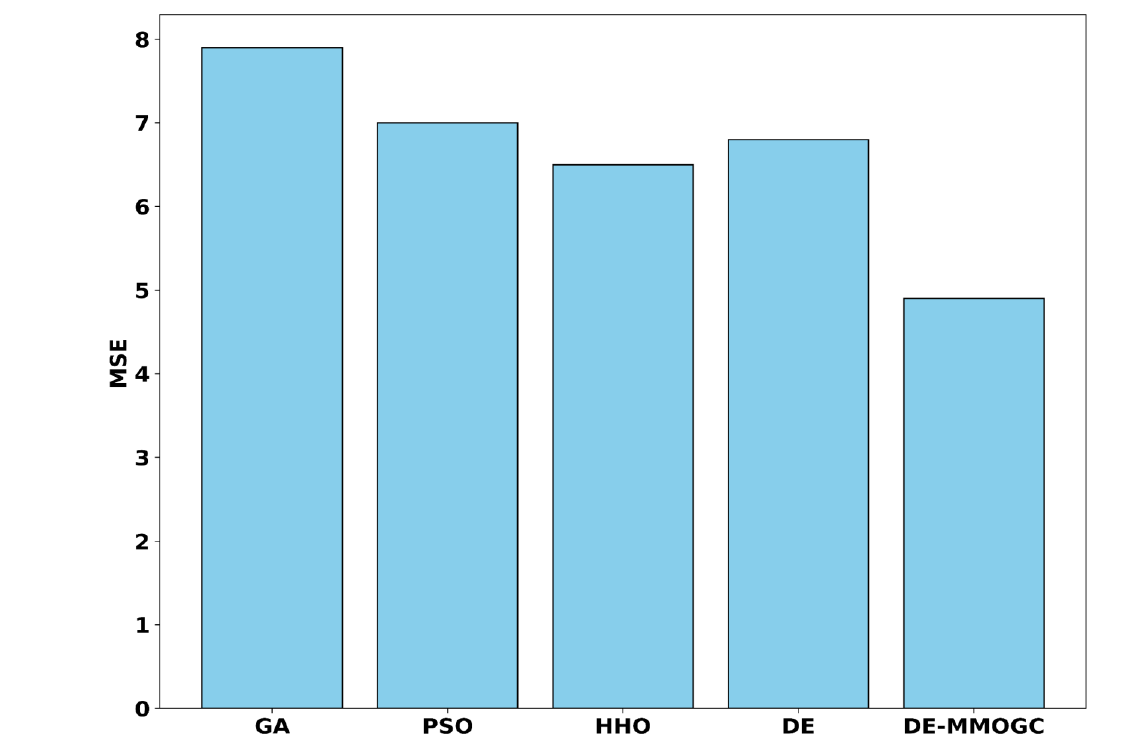}}
    \hfill
    \subcaptionbox{MAE}[0.48\linewidth]{\includegraphics[width=\linewidth,trim=0cm 16cm 1cm 1cm,clip]{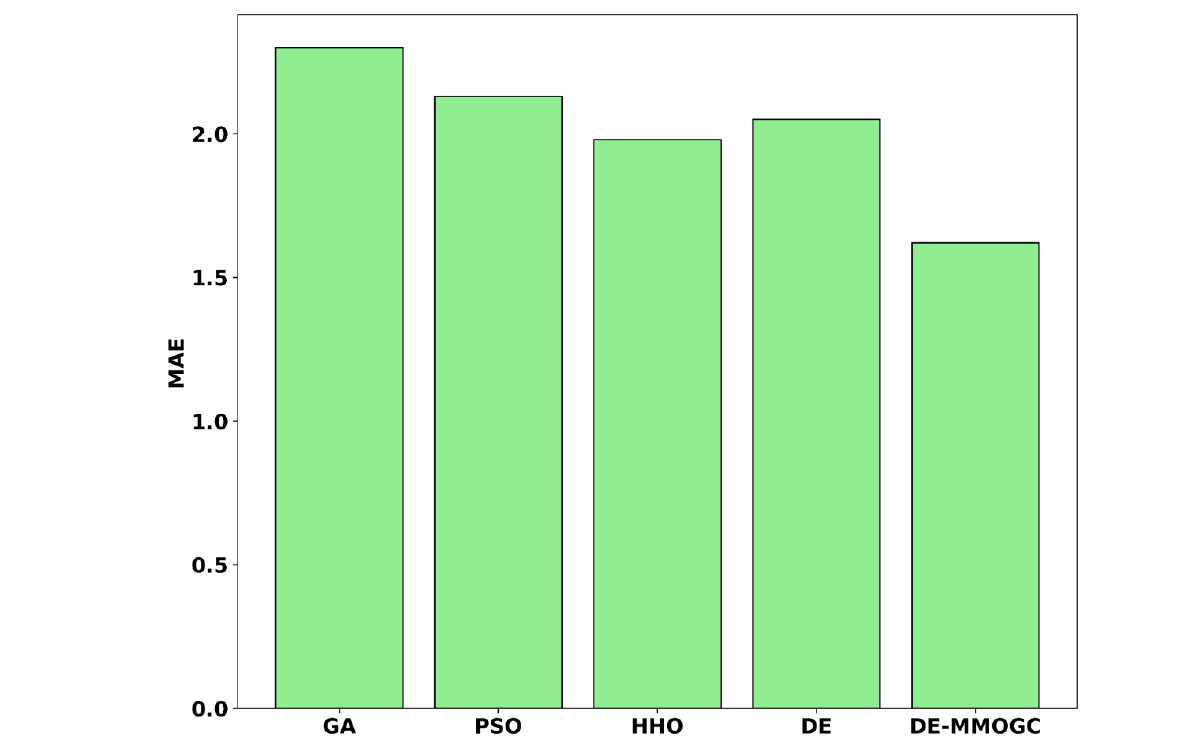}}
    
    \vspace{0.05cm}
    \subcaptionbox{RMSE}[0.48\linewidth]{\includegraphics[width=\linewidth,trim=0cm 16cm 2cm 1cm,clip]{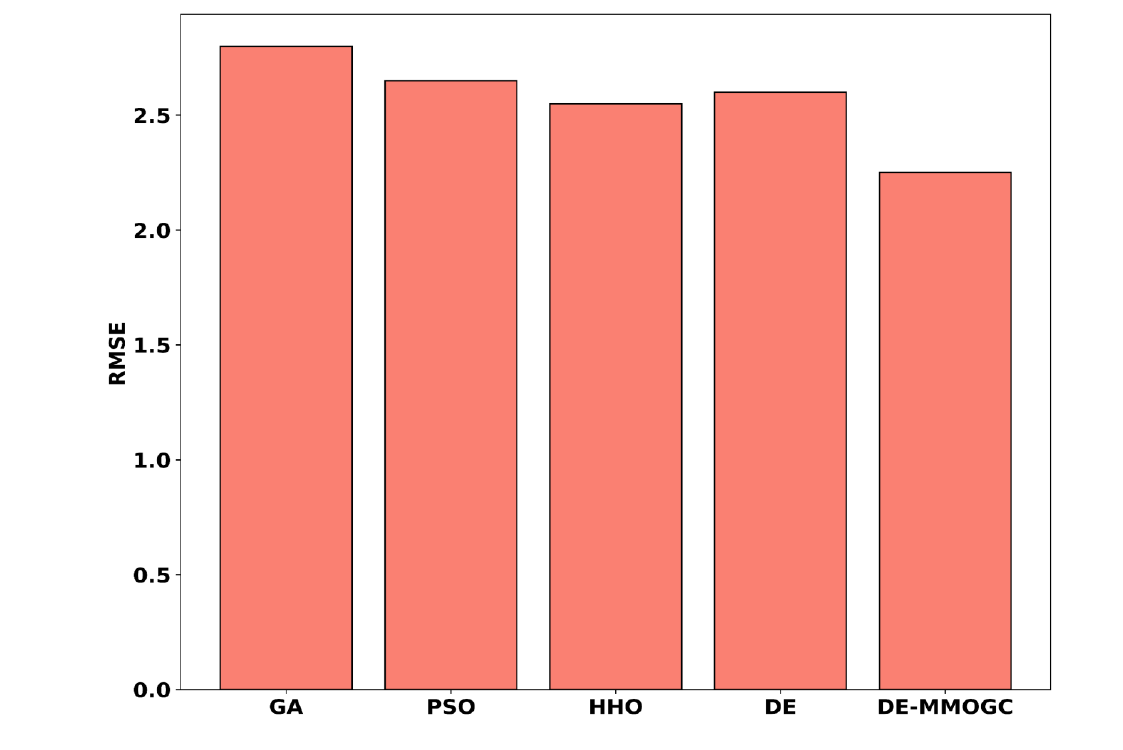}}
    \hfill
    \subcaptionbox{Correlation}[0.48\linewidth]{\includegraphics[width=\linewidth,trim=0cm 20cm 1cm 1cm,clip]{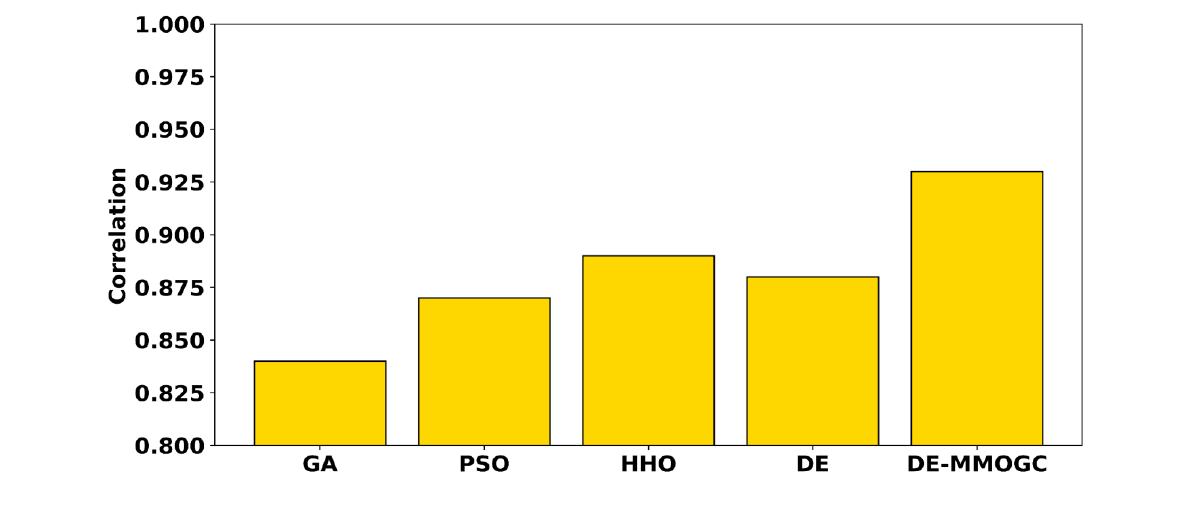}}   
    
    \caption{Performance comparison of DE-MMOGC and other optimization algorithms across four metrics. The figure highlights how DE-MMOGC achieves improved accuracy compared to its counterparts, demonstrating its superiority in handling complex model calibration problems.}
    
    \label{fig:graph1}
\end{figure*}

The comparison of four performance metrics across DE-MMOGC and alternatives is illustrated in Fig. \ref{fig:graph1} in the form of bar graphs. Taking MSE, MAE, and RMSE, the lower the number, the better the predictive accuracy. As we have shown in the graph, the proposed DE-MMOGC achieves the lowest error among all of these metrics. For the correlation metric, a higher value suggests better conformity to the true observation. By these graphs, we can validate the DE-MMOGC algorithm's higher efficiency in crop simulation model optimization.    

\par Further, as the comparative analysis demonstrates, single-mutation DE suffers from premature convergence and suboptimal exploration in high-dimensional, noisy, and dynamic environments like crop simulation models, despite its simplicity. The performance of GA and PSO is reasonable, but their flexibility in dynamic settings is limited. Because of the lack of population variety, HHO frequently stagnates in later iterations despite being competitive in initial convergence.

\par Conversely, we found that DE-MMOGC addresses these challenges effectively through three aspects: multi-mutation operator, communication-guided scheme, and dynamic operator selection. The multi-mutation system offers diversity, enabling the optimizer to maintain the equilibrium. The communication scheme guides the search process and hence avoids trapping in local minima. It also preserves the evolution across subpopulations. Finally, dynamic selection of operators ensures the improvement of mutation strategies based on their past performance. Based on their success probabilities, there is an adaptive use of mutation operators across subpopulations.

\subsection{Statistical Evaluation of DE-MMOGC}

\begin{table}[h]
    \centering
    \caption{Comprehensive statistical summary}
    \label{tab:stats}
    \begin{tabular}{p{0.20\linewidth} p{0.20\linewidth} p{0.20\linewidth} p{0.20\linewidth}}
      \toprule
       Algorithm  &  Mean & Median & Std \\
       \midrule
       \textbf{DE-MMOGC}  & \textbf{2.5859} & \textbf{2.5859} & \textbf{0.000605} \\
       GA & 2.9925 & 2.9979 & 0.019810 \\
       PSO & 2.7751 & 2.7750 & 0.000360 \\
       HHO & 2.9935 & 2.9935 & 0.000276 \\
       DE & 2.7675 & 2.7675 & 0.000609\\
       \bottomrule
    \end{tabular}
\end{table}

\begin{table}[h]
    \centering
    \caption{Wilcoxon rank-sum statistics}
    \label{tab:Wilcox}
    \begin{tabular}{p{0.30\linewidth} p{0.30\linewidth} p{0.20\linewidth}}
      \toprule
       Compared Algorithm  &  Wilcoxon Statistic & p-value\\
       \midrule
        GA & 0.0 & $2.0 \times 10^{-6}$\\
        PSO & 0.0 & $2.0 \times 10^{-6}$\\
        HHO & 0.0 & $2.0 \times 10^{-6}$\\
        DE & 0.0 & $1.0 \times 10^{-6}$\\
        \bottomrule
    \end{tabular}  
\end{table}

\begin{figure*}[h]
    \centering
    \includegraphics[width=\linewidth]{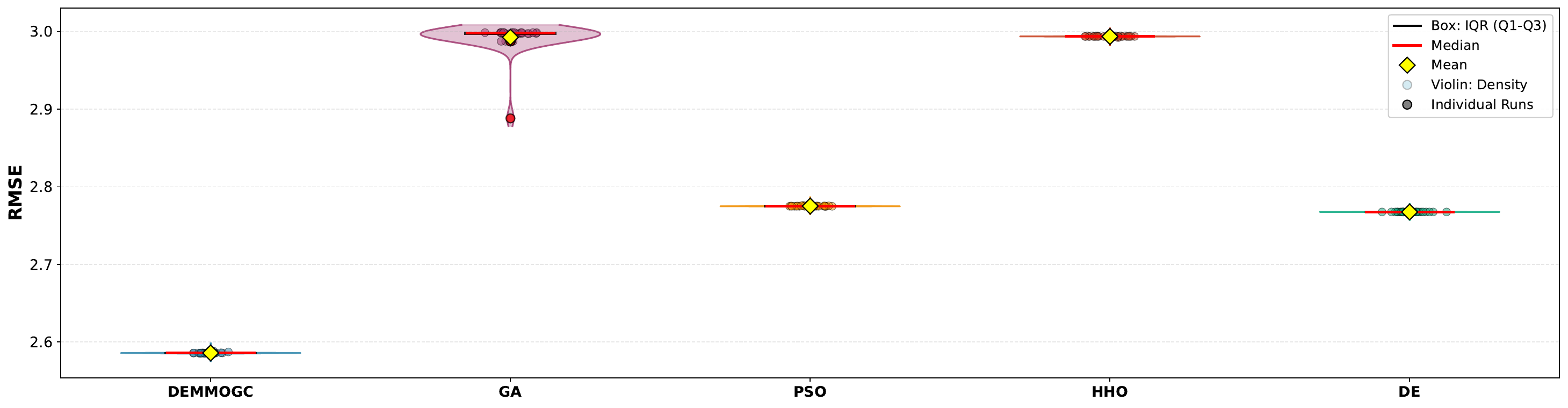}
    \caption{Box Plot with Violin Overlay and Swarm Points for Rice IR64 (Lowland) Variety}
    \label{fig:violinplot}
\end{figure*}

Over 30 separate runs, the proposed DE-MMOGC algorithm is assessed. The non-parametric Wilcoxon signed-rank test was used to determine statistical significance at the 5\% significance level. DE-MMOGC  beats all analysed algorithms, according to the statistical results. Table \ref{tab:stats} summarises the comprehensive statistics of the RMSE values extracted by each algorithm, including the mean, median, and standard deviation (std) for the Rice IR64 (Lowland) variety. 
\par With a mean RMSE value of 2.5859, the proposed DE-MMOGC algorithm outperforms all competing techniques by a significant range. GA and HHO, on the other hand, show the highest mean RMSE values of 2.9925 and 2.9935, respectively, suggesting lower-quality solutions. PSO and DE perform in an intermediate manner, with mean RMSE values of 2.7751 and 2.7675, respectively. DE-MMOGC stands out for having a very low standard deviation (0.000605), which shows stability throughout several separate runs. The mean RMSE values of PSO, HHO, and DE are nevertheless significantly greater than those of DE-MMOGC, even if they likewise exhibit minimal variability.
\par A combined violin-box plot of the RMSE distributions for each algorithm is shown in Fig. \ref{fig:violinplot}. With little dispersion and no extreme outliers, the DE-MMOGC distribution is tightly centred around its median, highlighting its dependable convergence properties. GA, on the other hand, shows greater performance variability with a larger distribution and noticeable spread. The central tendency of PSO and DE is still significantly lower than that of DE-MMOGC, despite their narrow distributions.
\par Table \ref{tab:Wilcox} presents the Wilcoxon rank-sum statistics and corresponding p-values for each pairwise comparison. The Wilcoxon rank-sum statistics of zero show that the RMSE values acquired by DE-MMOGC consistently rank lower than those generated by the competing algorithms across all separate runs. Instead of intermittent gains, this result shows a dominant and consistent performance advantage. The observed performance differences are highly statistically significant and unlikely to be the result of random fluctuations, as confirmed by the incredibly small p-values. DE-MMOGC retains a statistically better error profile even when compared to algorithms that show near-deterministic convergence behaviour (PSO, HHO, and DE).
\par The findings of the Wilcoxon rank-sum test, along with the distributional analysis and descriptive statistics, offer compelling statistical proof of DE-MMOGC's efficacy.

\subsection{Single-Mutation DE and Multi-Mutation DE}

\begin{table*}[!htbp]
    \centering
    \caption{Comparison table for single-mutation DE and multi-mutation-based DE-MMOGC for the Rice IR64 (Lowland) variety}
    \label{tab:singlevsmulti}
    \begin{tabular}{p{0.30\linewidth} p{0.10\linewidth} p{0.10\linewidth} p{0.10\linewidth} p{0.10\linewidth} }
    \hline
    Algorithm &  MSE &  MAE & RMSE & Correlation\\
     \hline 
     DE with mutation ``current-to-best/1'' & 6.7230  & 2.2478  & 3.1264  &  0.9251  \\
     DE with mutation ``rand-to-best/2'' & 7.2101 & 3.2458 & 3.1922 & 0.9760\\
     DE with mutation ``current-to-pbest/1'' &  6.7476 & 2.1563 & 2.6592 & 0.9753 \\
     \textbf{Proposed DE-MMOGC} & \textbf{6.6841} & \textbf{2.0085} & \textbf{2.5854} & \textbf{0.9814}\\  
    \hline
    \end{tabular}   
\end{table*}
Using the Rice IR64 (Lowland) variety, this section compares the individual mutation operators in the differential evolution assimilation strategy with the multi-mutation-based DE-MMOGC framework. Standard error measurements such as MSE, MAE, RMSE, and correlation coefficient are used to assess the performance in the Table \ref{tab:singlevsmulti}.
\par In every evaluation, the proposed DE-MMOGC consistently performs better than all single-mutation DE versions. In particular, DE-MMOGC obtains the highest correlation coefficient (0.9814) and the lowest MSE (6.6841), MAE (2.0085), and RMSE (2.5854). DE-MMOGC improves correlation and lowers RMSE by about 2.8\% when compared to the best-performing single-mutation variation (current-to-pbest/1), showing improved accuracy and higher alignment with ground data. These enhancements indicate that integrating several mutation techniques allows for more efficient search space exploration while maintaining exploitation capacity during convergence.
       
\section{Numerical Example of LAI prediction with EnKF Assimilation}
\label{appendix}
A numerical example of EnKF assimilation, demonstrating LAI prediction, is illustrated. We define the problem statement as:assimilating Leaf Area Index (LAI) observations into WOFOST using an optimized model with EnKF. The appendix includes all equations, step-by-step calculations, and explanations. 
\par Let's say the state variable $a_t$ at time $t$ is LAI. The observed LAI after some error is represented by $y_t$. Ensemble size $M$ is taken as $5$, i.e., $M=5$. Initially, true LAI, which is unknown in reality, is set to 0.6. All the notations and variables with their initial values are tabulated in Table \ref{tab:numerical}
\begin{table}[]
    \centering
    \caption{Variables used for EnKF numerical calculation}
    \label{tab:numerical}
    \begin{tabular}{p{0.15\linewidth} p{0.35\linewidth} p{0.35\linewidth}}
     \hline    
     Symbol    &  Description  &   Initial Value\\
    \hline
      $a_t$   &  State Vector: LAI & $[0.4\; \; 0.45\; \;0.5 \; \; 0.55 \;\;  0.6] $ \\
      $y_t$ & Observed LAI & $[0.48]$ \\
      $M$ & Ensemble Size & 5\\
      $R_t$ & Observation n oise & 0.01 \\
      $H_t$ & Maps model state to observations & 1\\
      $\mu$  & Mean & /  \\
      $P_t$  & Covariance & / \\
      $K_t$ & Kalman Gain & / \\
    \hline  
    \end{tabular}   
\end{table}

At day 1, the WOFOST simulation predicts the LAI value as,
\begin{ceqn}
\begin{align}
\label{observed}
\begin{split}
    a_{1} = WOFOST(a_0) = \begin{bmatrix}
        0.6 & 0.65 & 0.7 & 0.75 & 0.8
    \end{bmatrix}\\
    y_{1} = WOFOST(y_0) = [0.68]
\end{split}   
\end{align}
\end{ceqn}
Now, forecast mean $\mu$ is calculated as,
\begin{flalign}
\label{mean}
    &\mu = [0.6+ 0.65 + 0.7+ 0.75+ 0.8]/5  = 0.7&
    \end{flalign}
The covariance matrix $P_t$ in the EnKF measures the uncertainty in the model prediction and its relationship to observations. The calculation and application of covariance in the LAI assimilation example are broken out numerically below.
\begin{flalign}
     &  P_{t} = \frac{1}{M-1}\sum(a_{t}^i - \mu)^2 &
    \end{flalign} 

\begin{align}
    \begin{split}
        P_1 = \frac{1}{M-1} [(0.6-0.7)^2 + (0.65-0.7)^2 + (0.7-0.7)^2 +\\ (0.8-0.7)^2]
        P_1 = \frac{1}{4} (-0.1^2 + -0.05^2 + 0^2 +\\ 0.1^2)\\
        P_1 \approx 0.00625\\
    \end{split}      
    \end{align}

By using $P_1$, $R_t$, and $H_t$, compute Kalman Gain, $K_1$, that establishes a level of confidence in the model versus the observation, as given below,
    \begin{flalign}
    & K_t = P_{t}\mathcal{H}_t^T(\mathcal{H}_tP_{t}\mathcal{H}_t^T + R_t)^{-1} &   
    \end{flalign}

Since $H_t$ is a linear operator that maps directly to the observed LAI, hence, $H_t=1$. With the observation noise $R_t=0.01$, $K_1$ becomes,  
    \begin{flalign}
       & K_1 = \frac{0.00625}{0.00625 + 0.01} \approx 0.3846 & 
    \end{flalign}

Obtain the observed LAI perturbed with noise $\mathcal{N}(0, R_t)$. So $y_1$ is modified to
\begin{flalign}
      &  y_1 = 0.68 + \mathcal{N}(0,0.01) &
    \end{flalign}

For a simple discussion, let's say, $y_1$ is a perturbed vector having values, 
\begin{flalign}
     &  y_1 = \begin{bmatrix}
      0.72 & 0.67 & 0.68 & 0.65 & 0.70 
\end{bmatrix} &
    \end{flalign}

Finally, update each ensemble LAI value by EnKF assimilation, denoted by $a_{t+1}^E$, using the WOFOST crop growth model by the equation,
\begin{flalign}
      &  a_{t+1}^{E} = a_{t+1} + K_t(y_{t+1} - a_{t+1}) &
\end{flalign} 

    \begin{flalign}
        & a_1^E = \begin{bmatrix}
             0.6 + 0.3846(0.72-0.6)\\
             0.65 + 0.3846(0.67-0.65)\\
             0.7 + 0.3846(0.68-0.7)\\
             0.75 + 0.3846(0.65-0.75)\\
             0.8 + 0.3846(0.7-0.80)
         \end{bmatrix}^T  &
    \end{flalign}
\begin{flalign}
     &   a_1^E  = \begin{bmatrix}
            0.646 & 0.657 & 0.692 & 0.711 & 0.761
        \end{bmatrix} &
\end{flalign}
After getting the LAI value for day 1, $a_1^E$, the new mean, $\mu$, will be ,
    \begin{align}
    \begin{split}
        \mu = [0.6462+0.6577+0.6923+0.7115+0.7615]/5 \\
        \approx 0.6938
    \end{split}   
    \end{align}
The final results of EnKF assimilation and LAIs prediction by WOFOST and EnKF are recorded in Tables \ref{assim:result} and \ref{lais}. It is found that the updated mean \textit{0.6938} is pulled toward the observed LAI (as in equation \ref{observed}) \textit{0.68} compared to the forecast mean \textit{0.7}, from equation \ref{mean}. With the spread preserved, the ensemble is adjusted for observations. This verifies the theoretical nature of EnKF. 

\begin{table}[htbp]
    \centering
    \caption{LAIs values predicted with assimilation and no assimilation}
    \label{lais}
    \begin{tabular}{p{0.05\textwidth} | p{0.12\textwidth} | p{0.12\textwidth} | p{0.10\textwidth}}
      \hline
      Day  &  WOFOST Simulated LAI (No Assimilation) & EnKF Assimilated LAI & Observed LAI \\
      \hline
      1 & 0.7 & \textbf{0.6938} & 0.68\\
      \hline 
    \end{tabular}
\end{table}

\begin{table}[h]
    \centering
    \caption{EnKF Assimilation Results}
    \label{assim:result}
    \begin{tabular}{p{0.05\textwidth} | p{0.08\textwidth} | p{0.08\textwidth} | p{0.08\textwidth} | p{0.08\textwidth}}
     \hline
      Day   &  Forecast Mean $\mu$ & Forecast Covariance $P_t$ & Kalman Gain $K_t$ & Updated Mean\\
     \hline
      1 & 0.7 & 0.00625 & 0.3846 & \textbf{0.6938}\\
    \hline    
    \end{tabular}
\end{table}



\bibliographystyle{IEEEtran}
\bibliography{bibliography}